\documentclass{article} 
\usepackage{iclr2027_conference,times}

\usepackage{amsmath,amsfonts,bm}

\def\eqref#1{equation~\ref{#1}}

\def\1{\bm{1}}

\DeclareMathAlphabet{\mathsfit}{\encodingdefault}{\sfdefault}{m}{sl}
\SetMathAlphabet{\mathsfit}{bold}{\encodingdefault}{\sfdefault}{bx}{n}

\usepackage{hyperref}
\usepackage{xcolor}
\usepackage{longtable}
\usepackage{colortbl}
\usepackage{wrapfig}
\usepackage{amssymb}
\usepackage{capt-of}
\usepackage{float}
\definecolor{citeblue}{RGB}{0, 51, 153}
\definecolor{deepred}{RGB}{180, 25, 25}

\hypersetup{
    colorlinks=true,
    citecolor=citeblue,
    linkcolor=deepred,
    urlcolor=citeblue
}
\newcommand{\redeqref}[1]{%
    Eq.~\hyperref[#1]{\textcolor{deepred}{(\ref*{#1})}}%
}

\newcolumntype{L}[1]{>{\raggedright\arraybackslash}p{#1}}
\usepackage{array}
\usepackage{url}
\usepackage{enumitem}
\usepackage{booktabs}
\usepackage{multirow}
\usepackage{graphicx}
\usepackage{algorithm}
\usepackage{algpseudocode}
\usepackage{tabularx}
\usepackage{caption}

\title{OmniInteract: Towards Unified Whole-Body Interaction Control for Dual Humanoids}

\author{
\textbf{Yinghao Li}$^{1,2}$ \quad
\textbf{Tianyu Zhang}$^{2,3}$ \quad
\textbf{Zhuohao Li}$^{1,2}$ \quad
\textbf{Yi-Lin Wei}$^{1}$
\\
\textbf{Fa-Ting Hong}$^{4}$ \quad
\textbf{Qing Zhang}$^{1}$ \quad
\textbf{Wei-Shi Zheng}$^{1,2,\dagger}$
\\[1.5mm]
$^{1}$School of Computer Science and Engineering, Sun Yat-sen University
\\
$^{2}$Shenzhen Loop Area Institute
\qquad
$^{3}$Beijing Institute of Technology
\\
$^{4}$The Hong Kong University of Science and Technology
\\[0.8mm]
}

\iclrfinalcopy 
\begin{document}

\maketitle

\begingroup
\renewcommand{\thefootnote}{\fnsymbol{footnote}}
\footnotetext[2]{Corresponding author.}
\endgroup

\vspace{-8mm}
\lhead{Preprint}
\begin{figure}[!h]
    \centering
    \includegraphics[width=\linewidth]{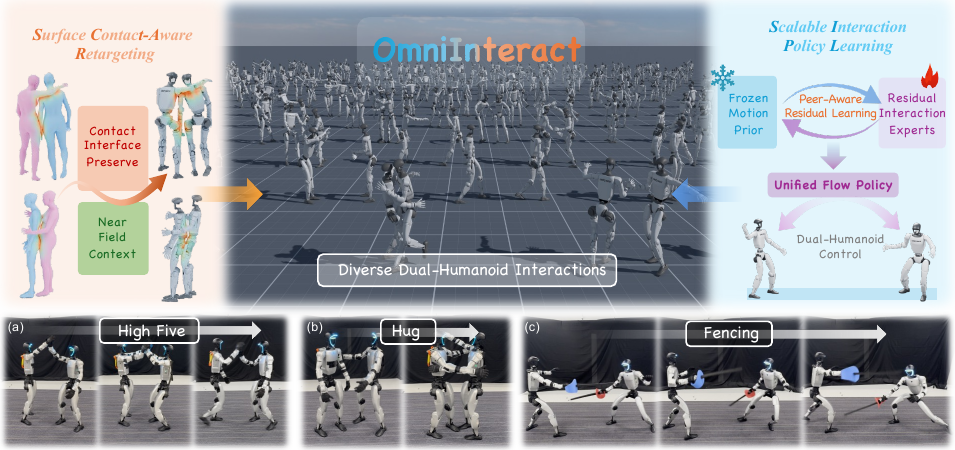}
    \captionof{figure}{
\textbf{Top center: OmniInteract enables diverse dual-humanoid interactions from human demonstrations.}
\textbf{Left:} SCAR (Surface Contact-Aware Retargeting) preserves fine-grained
surface interaction geometry during retargeting.
\textbf{Right:} Specialist-to-unified residual learning consolidates diverse
interaction skills into a unified flow policy.
\textbf{Bottom (Real, a--c):} Deployment on dual Unitree G1 robots performing
High-Five, Hug, and Fencing whole-body interactions.
    }
    \label{fig:o1}
\end{figure}

\begin{abstract}
Physical interaction is a key capability for humanoid collaboration, requiring
coordinated whole-body motion and precise contact. However, learning diverse dual-humanoid interactions faces two
major challenges: (1) human-robot morphological differences disrupt
contact interfaces and spatial relationships during
retargeting, and (2) jointly learning motor skills and partner coordination makes it difficult
to retain control fidelity as interaction diversity grows. In this paper, we propose \textbf{OmniInteract} to address these challenges
through surface-aware modeling and scalable residual policy learning. First, we introduce \textbf{Surface Contact-Aware Retargeting},
which leverages dense human-robot surface correspondences to jointly retarget
both humanoids, preserving explicit interaction contact interfaces and their
near-field geometric context. 
Second, we propose a \textbf{Scalable Flow-based Interaction Policy} for partner-conditioned residual generation. It factorizes dual-humanoid control into a reusable motor foundation and interaction-specific adaptations: partner-aware experts learn corrections over a frozen whole-body motion prior, while conditional flow matching unifies them in a shared residual action space. This enables scalable skill consolidation while retaining pretrained motor capabilities.
Extensive experiments demonstrate that OmniInteract improves
retargeting fidelity and dual-humanoid coordination while better retaining
interaction performance as skill diversity grows. Real-world deployment
further demonstrates successful contact establishment, stable whole-body
coordination, and recovery from external perturbations. \noindent\textbf{Project page:}
\href{https://calvinyhli.github.io/omniinteract.github.io/}{\textit{\textbf{https://calvinyhli.github.io/omniinteract.github.io/}}}
\end{abstract}
\section{Introduction}
Advances in whole-body control and motion imitation have equipped humanoid 
robots with increasingly versatile motor capabilities \citep{ben2025homie,fu2024humanplus,he2024omniho}. Most existing systems, however, focus on single-humanoid skills \citep{chen2025gmt,jiang2026wholebodyvla,he2025asap,li2026bfm}, whereas many real-world tasks require collaboration and physical
interaction between agents \citep{shaoul2025collaborative, doshi2026chorus}. Human-human interaction demonstrations offer a rich source of such behaviors.
Transferring them to dual humanoids requires preserving the intended
relationships between the bodies across embodiments
and maintaining coordinated motion and contact under coupled dynamics.

The first challenge is \textbf{\emph{preserving surface interaction geometry
during retargeting}.} Morphological differences in body proportions and surface shape
can disrupt contact interfaces despite plausible individual poses \citep{reda2023physics,yin2025dexteritygen}. Although recent interaction-aware approaches have begun to explicitly model relative geometries between bodies, sparse body representations remain insufficient for capturing fine-grained surface-level interactions \citep{chen2026rhythm}. Explicit contact relationships constrain where two bodies should interact, but they do not fully determine how surrounding limbs and body regions should be arranged \citep{liu2025takes}. Conversely, preserving neighborhood-level geometric structures may maintain overall spatial consistency while hiding local contact deviations. Therefore, High-fidelity retargeting requires preserving both where
the bodies meet and how nearby regions are arranged.

The second challenge is \textbf{\emph{retaining coordination fidelity as
interaction diversity grows}.} Independent motion trackers lack the
partner feedback needed to correct deviations under coupled dynamics.  Training from scratch instead requires learning individual tracking,
whole-body balance, and partner coordination together.
More importantly, different interactions exhibit
substantially different contact patterns and coordination requirements. As interaction diversity increases, jointly training multiple interaction behaviors within a unified policy can introduce severe
gradient interference, making it increasingly difficult to maintain high-fidelity interaction performance.

Motivated by this analysis, we propose \textbf{OmniInteract}, a unified
framework for learning diverse dual-humanoid interactions from human
demonstrations via surface-aware retargeting and scalable residual policy
learning, as illustrated in Fig.~\ref{fig:o1}. To overcome the retargeting bottleneck, we introduce \textbf{Surface Contact-Aware Retargeting (SCAR)}, which leverages dense human–robot surface correspondences to construct a contact-aware surface interaction
graph between humanoid. It preserves interaction geometry through complementary surface constraints: explicit bilateral contact edges maintain the contact interface, while near-field Laplacian relationships preserve the surrounding spatial structures. SCAR jointly optimizes both humanoids under
kinematic constraints, generating synchronized interaction motion references for downstream control.

To overcome the interaction control bottleneck, we propose a \textbf{Scalable Flow-based Interaction Policy} that consolidates partner-dependent corrections from task-specific dual-humanoid experts into a unified generative residual controller through flow matching. This design decouples stable single-humanoid motion priors from interaction control learning, leveraging high-capacity generative policies to expand interaction representation capability while avoiding severe interference multi-interaction behaviors reinforcement learning. Extensive experiments demonstrate that OmniInteract improves
retargeting fidelity and dual-humanoid coordination, together with stronger
generalization to unseen motions and better performance retention as
interaction diversity grows. Deployment on two Unitree G1 robots
further demonstrates successful contact establishment, stable
whole-body coordination, and recovery from external pushes. Our contributions are summarized as follows:

\begin{itemize}[leftmargin=*]
    \item We present \textbf{OmniInteract}, a unified framework for
    learning dual-humanoid interactions from human demonstrations,
    integrating interaction-preserving retargeting with scalable
    coordination.

\item We introduce a \textbf{surface contact-aware retargeting (SCAR)} method that leverages dense human--robot surface correspondences to preserve bilateral contact interfaces and surrounding interaction geometry for generating synchronized dual-humanoid references.
\item We propose a \textbf{scalable flow-based interaction policy}  that distills partner-aware corrections into a unified residual action space, enabling diverse coordination control while retaining a stable whole-body motion prior.
\end{itemize}
\section{Related Works}

\paragraph{Motion Retargeting for Humanoid.}
Retargeting human motions to humanoid robots is a fundamental step for transforming large-scale human demonstrations into robot-executable references \citep{park2025demodiffusion,luo2023perpetual}. 
Early approaches typically rely on manually designed correspondences between human and robot joints or body parts, followed by optimization under kinematic constraints \citep{araujo2025retargeting,qin2022one}. Recent studies have improved retargeting quality by incorporating whole-body geometric structures, interaction constraints, and physical feasibility \citep{yang2025omniretarget,pan2025spider}. However, existing representations often rely on sparse body configurations or predefined correspondences, which is insufficient to capture fine-grained surface interactions between multiple humanoids \citep{chen2026rhythm,liu2025takes}. 
Recent dense correspondence-based approaches represent humanoid bodies through surface-level geometric correspondence \citep{cao2026unified}, enabling more scalable motion transfer across embodiments. Our work builds upon this direction and extends surface-based retargeting toward dual-humanoid interaction by explicitly modeling bilateral contact interfaces and surrounding interaction geometry.

\paragraph{Whole-body Control for Humanoid Robots.}
Learning robust whole-body control policies for humanoid robots has recently advanced through motion imitation \citep{ji2024exbody2,xie2026kungfubot,han2025kungfubot2}, reinforcement learning \citep{li2026w1, peng2018deepmimic,zhang2025falcon}, and generative policy learning \citep{li2026language, liu2025commanding, shao2025langwbc}. Prior works have demonstrated impressive capabilities in single-humanoid locomotion, manipulation \citep{gu2026humanoid, he2026viral}, and dynamic motion tracking \citep{wang2026omnixtreme}. Extending these abilities to dual-humanoid interaction introduces additional challenges, as the policy must simultaneously maintain individual motion stability and adapt to partner-dependent coordination requirements. Existing dual-humanoid control approaches typically learn interaction policies directly through reinforcement learning or task-specific optimization, which can become difficult to scale as interaction diversity increases \citep{liu2025takes,chen2026rhythm}. 
In contrast, our approach leverages pretrained single-humanoid motion priors and learns interaction-specific residual adaptations, followed by specialist-to-unified policy consolidation, enabling scalable learning of diverse dual-humanoid interactions while preserving stable individual motion capabilities.
\section{OmniInteract}
\label{sec:omniinteract}
\begin{figure}[t]
    \centering
    \includegraphics[width=\linewidth]{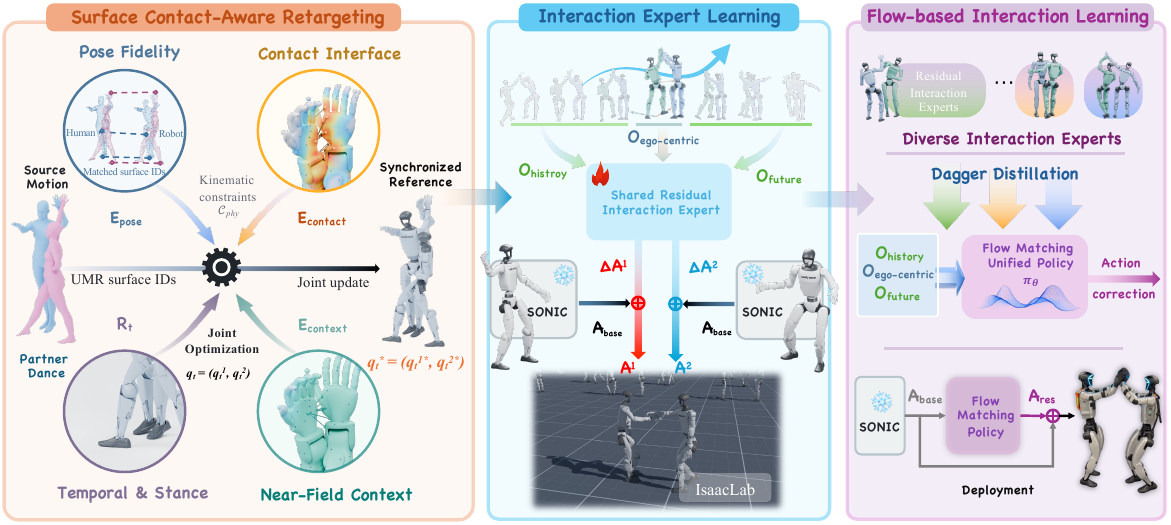}
\caption{
\textbf{Overview of the \textsc{OmniInteract} pipeline.}
\textbf{Left:} Starting from human-human demonstrations, SCAR jointly retargets the two
humanoids while preserving individual motion, surface contacts, and near-field
interaction context, yielding synchronized inetraction motions. \textbf{Center:} A frozen SONIC
tracker provides the base actions, while task-specific residual experts learn
partner-aware interaction corrections. \textbf{Right:} These diverse interaction experts are then distilled into a unified flow matching policy through DAgger. During deployment, the unified policy predicts residual actions on top of the frozen SONIC for coordinated dual-humanoid control.
}
    \label{fig:overview}
    \vspace{-15pt}
\end{figure}

Our goal is to learn whole-body humanoid--humanoid interactions from human--human demonstrations. Unlike independent motion imitation, this task requires preserving inter-body geometry across embodiments during retargeting and sustaining coordinated behavior under coupled dynamics during control.
Therefore, we present \textbf{OmniInteract}, a unified framework combines \textbf{surface contact-aware retargeting (SCAR)} with \textbf{scalable flow-based interaction policy training} to address these challenges. In this section, we detail (i) how SCAR jointly retargets dual humanoids while preserving contact surface and surrounding spatial geometry, and (ii) how our flow-based interaction policy training specializes partner-aware residual corrections and consolidates them into a unified interaction policy. Fig.~\ref{fig:overview} illustrates the overall pipeline of our framework.

\subsection{Surface Contact-Aware Retargeting}
\label{sec:scar}

\paragraph{Challenge of Interaction Retargeting.}
Conventional motion retargeting methods typically rely on sparse skeletal or
keypoint constraints, which preserve coarse body motion but provide insufficient
supervision for the surface contacts that define close human interactions. For example, in an embrace, aligning the wrists and torso does not ensure that
the palms reach the intended regions of the partner's back. The core challenge lies in synthesizing motions that simultaneously preserve precise inter-body surface contacts and their surrounding local geometry.


\paragraph{Surface Correspondence and Interaction Coordinates.}
To model fine-grained inter-body surface contacts, we adopt Dense Point Cloud Correspondence \citep{cao2026unified} to establish surface-level associations across embodiments. For humanoid $a\in\{1,2\}$, let $\mathit h_{t,i}^{a}$ denote the
$i$-th source surface point at time $t$, and
$\mathit{x}_{i}^{a}(q_{t}^{a})$ its corresponding robot surface point under
configuration $q_{t}^{a}$. The correspondence is established in the canonical pose and remains fixed throughout the motion,while each
robot point is updated through forward kinematics according to its local body binding. See Appendix~\ref{app:scar_source} for preprocessing and surface-binding details.

To represent how a surface point is arranged relative to the opposite body, we utilize Laplacian coordinates \citep{alexa2003differential,zhou2005large} to define cross-body interaction coordinates. Specifically, For a surface point $\mathit p_i$ (human or robot), let $\mathcal N_t(i)$ denote its local
neighbors on the opposite body.
The interaction surface coordinates can be formulated as:
\begin{equation}
\mathcal L_t(\mathit p_i)
=
\mathit p_i
-
\sum_{j\in\mathcal N_t(i)}
\omega_{t,ij}\mathit p_j.
\label{eq:surface_laplacian}
\end{equation}
Here, $\omega_{t,ij}$ denotes the normalized weight of neighbor $j$,
computed from the source geometry. $\mathcal L_t(\mathit p_i)$ encodes the local spatial relation between a surface point and the opposite body. Minimizing the deformation energy can therefore preserves local inter-body geometry across embodiments.

\paragraph{Optimization Objective.}
Given the surface correspondences and interaction coordinates defined above,
SCAR jointly retargets the two humanoids by balancing individual motion
fidelity with inter-body interaction preservation. Let
$q_t=(q_t^{1},q_t^{2})$ denote the root poses and joint configurations of
the two robots at time $t$. We optimize the paired configuration through
the following objective:
\begin{equation}
q_t^\star
=
\underset{q_t\in\mathcal C_{\mathit{phy}}}{\operatorname{arg\,min}}
\;
\mathcal E_{\mathit{pose}}(q_t)
+\lambda_{\mathit{c}}\mathcal E_{\mathit{contact}}(q_t)
+\lambda_{\mathit{n}}\mathcal E_{\mathit{context}}(q_t)
+\mathcal R_t(q_t).
\label{eq:scar_objective}
\end{equation}
Here, $\mathcal E_{\mathit{pose}}$ preserves individual motion by matching corresponding surface positions and orientations, $\mathcal E_{\mathit{contact}}$ preserves explicit contact interfaces, and $\mathcal E_{\mathit{context}}$
maintains the surrounding near-field geometry. $\mathcal R_t$ regularizes temporal motion and stance-foot
drift, and $\mathcal C_{\mathit{phy}}$ defines the feasible set under joint,
velocity, floor, and collision constraints. The two humanoids are optimized jointly in a coupled configuration space. We solve \redeqref{eq:scar_objective} sequentially for each timestep using a Sequential Quadratic Programming (SQP) approach, warm-started with the optimal solution from the previous frame $q_t^\star$. Solver details, constraint handling, and hyperparameters are provided in Appendix~\ref{app:scar_optimization}.

\paragraph{Contact-Interface Objective.}
To explicitly preserve physical contact at the surface level, we construct
a sampled contact interface from the dense source correspondences. At each frame, we perform bidirectional nearest-neighbor search
between the two source surfaces and collect pairs whose distances fall within
a contact threshold. Temporal hysteresis is then applied to stabilize persistent
contacts, followed by region-balanced subsampling to obtain a budgeted
contact set $\mathcal C_t$. For each $(i,j)\in\mathcal C_t$, we define the demonstrated and retargeted
surface offsets as
$\mathit d_{t,ij}^{h}=\mathit h_{t,i}^{1}-\mathit h_{t,j}^{2}$ and
$\mathit d_{t,ij}^{r}(q_t)=
\mathit x_i^{1}(q_t^{1})-\mathit x_j^{2}(q_t^{2})$, respectively.
We preserve the contact interface by minimizing
\begin{equation}
\mathcal E_{\mathit{contact}}(q_t)
=
\sum_{(i,j)\in\mathcal C_t}
\alpha_{t,ij}
\left\|
\mathit d_{t,ij}^{r}(q_t)
-
\mathit d_{t,ij}^{h}
\right\|_2^2,
\label{eq:scar_contact}
\end{equation}
where $\alpha_{t,ij}$ is the normalized weights from the source
surface. This objective preserves both the distance and direction of each contact interface by matching the relative contact-offset vectors.

\paragraph{Near-Field Context Objective.}
To maintain the arrangement of nearby limbs and surfaces, we construct a
compact set of near-field context relations around the contact interface.
For source surface points that were not selected as contact endpoints, we first
identify near-field candidates according to their distance to the opposite
body.
For each humanoid $a\in\{1,2\}$, we retain the closest candidates up to
a fixed number to form the context-center set $\mathcal S_t^{a}$.
For each center $i\in\mathcal S_t^{a}$, we find its nearest point on the opposite
body and augment it with valid one-ring surface neighbors within a local
distance threshold to form $\mathcal N_t(i)$. Using the interaction
coordinates defined in \redeqref{eq:surface_laplacian}, we minimize
\begin{equation}
\mathcal E_{\mathit{context}}(q_t)
=
\sum_{a\in\{1,2\}}
\sum_{i\in\mathcal S_t^{a}}
\beta_{t,i}^{a}
\left\|
\mathcal L_t\!\left(\mathit x_i^{a}(q_t^{a})\right)
-
\mathcal L_t\!\left(\mathit h_{t,i}^{a}\right)
\right\|_2^2 .
\label{eq:scar_context}
\end{equation}
where $\beta^a_{t,i}$ is the normalized
proximity weight of context center $i$.
This objective aims to preserve the local organization of near-field surface structure while accommodating morphological differences.
Contact selection, neighborhood construction, and weight definitions are detailed in Appendix~\ref{app:scar_optimization}.

\subsection{Scalable Flow-Based Interaction Policy Training}
\label{sec:interaction_policy}

\paragraph{Problem Formulation and Challenge of Interactive Control.}
We formulate the interaction control as a Multi-Agent Markov Decision Process (MA-MDP):
$\mathcal G=(\mathcal S,\ \mathcal O, \mathcal A,\mathcal P, \mathcal R,\gamma)$, where $\mathcal S$ is the global state space, $\mathcal O=\prod_{a}\mathcal O^a$ and
$\mathcal A=\prod_{a}\mathcal A^a$ are the joint observation and action
spaces of humanoid $a\in\{1,2\}$, while the transition function $\mathcal{P}:\mathcal{S}\times\mathcal{A}\rightarrow\mathcal{S}$
specifies the environment dynamics. The reward function
$\mathcal{R}:\mathcal{S}\times\mathcal{A}\rightarrow\mathbb{R}$
provides a shared interaction scalar signal, and $\gamma$ is the discount factor. The objective is to learn a joint policy
$\pi$ that maximizes $\mathbb E_{\pi}[\sum_{\tau=t}^{\infty}\gamma^\tau r_\tau]$.

Given the interaction motions $\mathcal{M}^a$ ($a\in\{1,2\}$) retargeted by SCAR, our goal is to achieve
coordinated execution under coupled dynamics. However, most existing
strong trackers are designed for single-humanoid and cannot
effectively handle partner-dependent deviations induced by physical
interaction. Training a dual-humanoid controller from scratch, in contrast,
requires simultaneously learning motion tracking, balance, and inter-body
coordination. This challenge becomes more pronounced as increasingly diverse interaction
patterns are incorporated into a single policy.

\paragraph{Residual Interaction Expert Learning.}
To avoid relearning dual-humanoid low-level motor control from
scratch, we build upon
\textit{SONIC} \citep{luo2026sonic}, a strong general whole-body tracker, and preserve its
pretrained tracking and balance capabilities. We freeze the SONIC policy
$\pi_{\mathit{base}}$ and learn only the interaction-specific correction
through a residual expert $\pi^k_{\textit{res}}$ for each interaction task $k$. For humanoid $a$ and its partner $\bar{a}$, the final corrected action is formulated as:
\begin{equation}
{\mathcal A}^a
=
\underbrace{
\pi_{\mathit{base}}(\mathcal O^a,\mathcal M^a)
}_{\textcolor{deepred}{
\text{frozen action}\;
\mathcal A_{\mathit{base}}^a
}}
+
\underbrace{
\alpha_{\mathit{res}}
\tanh\!\bigl(
\pi^k_{\textit{res}}(\mathcal O^a_{his},\mathcal O^a_{ego},\mathcal O^a_{fur})
\bigr)
}_{\textcolor{deepred}{
\text{residual interaction action}\;
\Delta\mathcal A^{\bar{a}\rightarrow a}
}}.
\label{eq:interaction_residual}
\end{equation}
Here, $\mathcal A_{\mathit{base}}^a$ denotes the base action produced by
the frozen SONIC tracker, while $\Delta\mathcal A^{\bar a\rightarrow a}$
represents the partner-dependent residual correction for humanoid $a$. $\alpha_{\mathit{res}}$ bounds the residual magnitude.
Here, $\mathcal O^a_{his}$ encodes the recent history of self and partner
states, previous actions, and relative-geometry errors;
$\mathcal O^a_{ego}$ represents the partner's state and relative pose
in the ego-centric frame; and $\mathcal O^a_{fur}$ provides short-horizon
motion references for both humanoids and their desired relative geometry. Both are encoded by temporal 1D CNNs, and the resulting features are
fused through an MLP decoder to predict the residual action.
For each interaction task, the residual expert is shared by both humanoids
and trained with Multi-Agent Proximal Policy Optimization (MAPPO) under centralized training with decentralized execution \citep{littman1994markov, gao2024coohoi}, using a shared reward composed of motion tracking, interaction-geometry
alignment, contact-timing consistency, and residual regularization.
This design decouples interaction coordination from low-level tracking and
balance, allowing reinforcement learning to focus on partner-dependent
corrections rather than relearning individual control. Detailed observations, network architecture and reward definitions are provided in
Appendix~\ref{app:sipl_experts}.

\paragraph{Flow-Matching Interaction Policy Learning.}
To consolidate task-specific interaction skills into a unified policy, we
distill the residual experts into a flow-based interaction policy using
DAgger \citep{ross2011reduction}. We first roll out the current flow-based policy $\pi_\theta(\Delta \mathcal{A}\mid \hat{\mathcal{O}})$ on top of the frozen
Sonic base action in dual humanoids simulation and collecting a trajectory of visited states given interaction motions dataset retargeted by SCAR, where $\hat{\mathcal O}$ contains the motion histories and future reference
features of both humanoids. For each visited state, we obtain the residual interaction expert action $\Delta \mathcal{A}_\mathrm{expert}$ by querying the corresponding interaction expert policy. Since the Sonic tracker remains frozen and
shared across all experts, all supervision lies in a common residual action
space, allowing the unified policy to focus on interaction coordination
without relearning individual motion. The flow-based model then learn to recover the interaction expert residual action $\Delta \mathcal{A}_\text{expert}$ from noised action by optimizing:
\begin{equation}
\mathcal L_{\mathit{fm}}(\theta)
=
\mathbb E_{t,\epsilon,\Delta \mathcal{A}_{\mathrm{expert}}}\!\left[
\left\|
v_\theta(\Delta \mathcal{A}_t,t,\hat{\mathcal O})
-
(\boldsymbol\epsilon - \Delta \mathcal{A}_\mathrm{expert})
\right\|_2^2
\right],
\label{eq:interaction_flow}
\end{equation}
where $\Delta\mathcal A_t=(1-t)\Delta\mathcal A_{\mathrm{expert}}
+t\boldsymbol\epsilon$ is the noisy residual action interpolated between the
expert residual $\Delta\mathcal A_{\mathrm{expert}}$ and Gaussian noise
$\boldsymbol\epsilon\sim\mathcal N(\mathbf 0,\mathbf I)$ along the flow
timestep $t\in[0,1]$ \citep{lipman2023flow}. This objective
learns a conditional vector field $v_\theta(\Delta\mathcal A_t,t,\hat{\mathcal
O})$ to predict the target velocity
$\boldsymbol\epsilon-\Delta\mathcal A_{\mathrm{expert}}$, learning the
denoising direction from expert residual actions to random noise along the
flow trajectory.
With the learned vector field, the residual interaction action can be
generated from Gaussian noise by integrating $v_\theta$ from $t=1$ to $t=0$
using the forward Euler rule:
\begin{equation}
\Delta\mathcal A_{t-\frac{1}{N}}
=
\Delta\mathcal A_t
-
\frac{1}{N}
v_\theta(\Delta\mathcal A_t,t,\hat{\mathcal O}),
\end{equation}
where $N$ denotes the number of integration steps controlling the sampling
accuracy. By iteratively rolling out trajectories and supervising them with
interaction expert residuals through DAgger, we learn a unified flow-based
policy that maps interaction conditions to partner-aware residual
corrections. During deployment, the generated residual action is added to the
frozen Sonic base action for dual-humanoid control.
Detailed deployment settings are provided in the Appendix~\ref{app:deploy}.
\section{Experiments}
We present extensive experiments in simulation and on physical robots to evaluate the interaction retargeting fidelity, control scalability, and real-world executability of our proposed \textbf{OmniInteract} framework. Our experiments are organized around the following key questions:

\begin{itemize}[leftmargin=*]

\item \textbf{Q1: Interaction-Preserving Retargeting.}
Can \textbf{SCAR} generate feasible trajectories that better preserve
fine-grained inter-body geometry than prior retargeting methods? (Sec. ~\ref{q1})

\item \textbf{Q2: Residual Interaction Learning.}
Do residual experts built on frozen motion priors improve dual-humanoid
coordination over independent or scratch-trained controllers? (Sec. ~\ref{q2})

\item \textbf{Q3: Flow-Based Interaction Scaling.}
Does scaling a flow-based generative policy improve multi-interaction control
more effectively than conventional MLP controllers? (Sec. ~\ref{q3})

\item \textbf{Q4: Real-World Robustness.}
Can \textbf{OmniInteract} transfer to physical humanoids under perturbations
while maintaining accurate contacts and stable whole-body coordination? (Sec. ~\ref{q4})

\end{itemize}
\subsection{Experiments settings}
\paragraph{Datasets.} We evaluate OmniInteract on two human--human interaction datasets, \textbf{InterHuman}~\citep{liang2024intergen} and
\textbf{Inter-X}~\citep{xu2024inter}. We use the full motion sets from both datasets for   interaction retargeting evaluation. For control, we select 30 diverse interaction motions from each dataset to form the control benchmark to assess performance across physical dynamics. Moreover, we further construct a unseen motion set to evaluate generalization beyond the curated interaction motion library. More dataset details are provided in Appendix~\ref{app:bench}.
\vspace{-5pt}

\paragraph{Baselines.} We compare against representative baselines for both interaction retargeting
and control.
For retargeting, we evaluate
PHC~\citep{luo2023perpetual},
GMR~\citep{araujo2025retargeting},
OmniRetarget~\citep{yang2025omniretarget}, and
UMR~\citep{cao2026unified}
using the same source motions and skeletal-scaling preprocessing.
For control, we consider two groups.
\emph{Single-humanoid trackers} include
TWIST2~\citep{ze2025twist2},
BeyondMimic~\citep{liao2026beyondmimic}, and
SONIC~\citep{luo2026sonic}, which independently execute the paired
reference motions for dual humanoids.
\emph{Peer-aware variants} include
\textbf{From-scratch RL}, which directly learns full-body control without
frozon motion priors, and
\textbf{Specialist$\rightarrow$Unified MLP}, which retains the residual
specialists but replaces flow matching with deterministic regression.
These comparisons isolate the advantages of residual interaction learning
and flow-based policy consolidation. Implementation for all baselines are detailed in Appendix~\ref{app:baseline}.

\paragraph{Metrics.}
We evaluate retargeting quality and control performance using task-specific metrics. For retargeting, we report
\textbf{MPD}(cm) for physical plausibility,
\textbf{CIE}(\%), \textbf{NGE}(\%), and
\textbf{C-F1}(\%) for interaction fidelity, and
\textbf{DSR}(\%) for downstream learnability.
For control, we report
$E_{\textit{mpjpe}}$(mm) for motion tracking error,
$E_{\textit{inter-geo}}$(\%) for interaction preservation,
and \textbf{Succ}(\%) for overall interaction execution success. For conciseness, detailed metric definitions are provided in
Appendix~\ref{app:metrics}.

\begin{table}[t]
\centering
\caption{\textbf{Quantitative Results of Retargeting.} Metrics include physical feasibility (MPD), interaction fidelity (CIE, NGE, C-F1), and utility (DSR). Best results are shown in
\textbf{bold}.
}
\label{tab:retargeting}

\small
\renewcommand{\arraystretch}{1.05}

\begin{tabular*}{\linewidth}{
@{\extracolsep{\fill}}
llccccc
@{}
}
\toprule
\multirow{2}{*}{\textbf{Dataset}}
& \multirow{2}{*}{\textbf{Method}}
& \multicolumn{1}{c}{\textbf{Feasibility}}
& \multicolumn{3}{c}{\textbf{Interaction Fidelity}}
& \multicolumn{1}{c}{\textbf{Utility}} \\
\cmidrule(lr){3-3}
\cmidrule(lr){4-6}
\cmidrule(lr){7-7}

&
& MPD$_{(\mathrm{cm})}\downarrow$
& CIE$_{(\%)}\downarrow$
& NGE$_{(\%)}\downarrow$
& C-F1$_{(\%)}\uparrow$
& DSR$_{(\%)}\uparrow$ \\
\midrule

\multirow{5}{*}{\textit{InterHuman}}
& PHC
& \underline{2.59}
& 10.07
& 10.89
& 47.32
& 40.6 \\

& GMR
& 4.76
& 13.94
& 13.03
& 31.98
& 36.4 \\

& OmniRetarget
& 3.54
& \underline{6.71}
& \underline{6.91}
& \underline{58.69}
& \underline{58.7} \\

& UMR
& 4.04
& 7.83
& 7.73
& 51.14
& 56.9 \\

& \textbf{SCAR (Ours)}
& \textbf{0.93}
& \textbf{4.96}
& \textbf{6.49}
& \textbf{65.18}
& \textbf{75.4} \\

\midrule

\multirow{5}{*}{\textit{Inter-X}}
& PHC
& \underline{3.15}
& 10.94
& 11.12
& 33.90
& 31.2 \\

& GMR
& 5.10
& 9.64
& 9.36
& 43.07
& 34.6 \\

& OmniRetarget
& 5.02
& 8.09
& 8.32
& \underline{56.21}
& 48.7 \\

& UMR
& 4.98
& \underline{7.19}
& \underline{7.04}
& 50.03
& \underline{55.1} \\

& \textbf{SCAR (Ours)}
& \textbf{1.20}
& \textbf{3.98}
& \textbf{4.99}
& \textbf{64.08}
& \textbf{72.1} \\

\bottomrule
\end{tabular*}
\vspace{-5pt}
\end{table}

\begin{figure}[t]
    \centering
    \includegraphics[width=\linewidth]{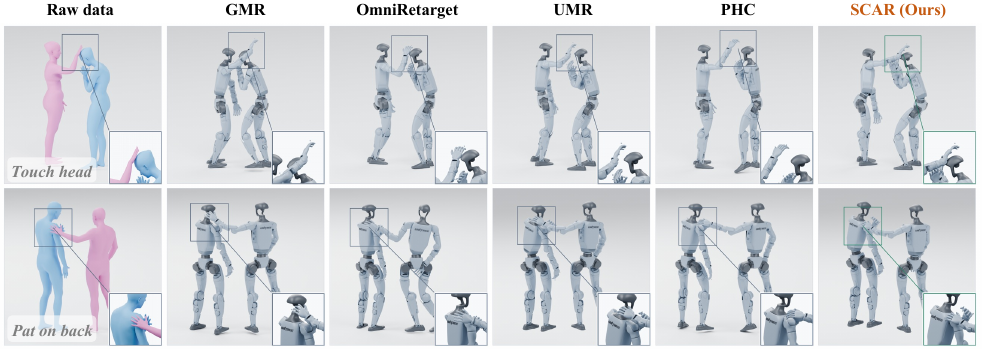}
    \caption{Qualitative visualization comparison of interaction retargeting.
(\textbf{Top}: Touch Head, \textbf{Bottom}: Pat on Back)}
    \label{fig:retarget_vis}
    \vspace{-15pt}
\end{figure}

\subsection{Fidelity: Interaction-Preserving Retargeting (Q1)}
\label{q1}
\paragraph{Quantitative Analysis.}
Table~\ref{tab:retargeting} reveals three clear trends.
First, sparse single-body fitting is insufficient for interaction preservation:
GMR exhibits consistently large CIE and NGE, indicating that accurate keypoint
tracking alone cannot recover fine-grained cross-body geometry.
Second, surface-aware baselines improve interaction fidelity but still struggle
to balance geometry and feasibility when the two humanoids are optimized
independently. OmniRetarget suffers from noticeable penetration and degraded
near-field consistency, while UMR benefits from dense surface correspondences
but still lacks explicit pairwise interaction constraints.
Third, SCAR consistently resolves this trade-off, achieving the best MPD, CIE,
NGE, and C-F1 on both datasets. On Inter-X, it reduces CIE and NGE by $44.6\%$
and $29.1\%$ over the second-best methods. Importantly, this improvement also
translates to downstream learnability, with DSR reaching $75.4\%$ and $72.1\%$
on InterHuman and Inter-X. These results highlight that faithful interaction
retargeting requires modeling geometry between the two bodies, rather
than fitting each body independently.

\paragraph{Qualitative Analysis.}
Fig.~\ref{fig:retarget_vis} compares the retargeted interactions with the
source motions, with close-up views highlighting surface alignment.
In \emph{Touch Head} (top, InterHuman), baselines reproduce the overall
gesture but exhibit displaced or separated hand--head contact, whereas
SCAR better preserves the intended contact location. In \emph{Pat on Back}
(bottom, Inter-X), SCAR more faithfully retains the hand--back interface
and the relative arm--torso arrangement. These examples highlight the
importance of preserving contact interface and near-field context.

\begin{table}[t]
\centering
\caption{\textbf{Comparison of controllers on dual-humanoid interaction benchmarks.}
InterHuman$^{\dagger}$ and Inter-X$^{\dagger}$ denote the motion subsets selected from
each dataset for control evaluation.}
\label{tab:control}

\small
\setlength{\tabcolsep}{4.5pt}
\renewcommand{\arraystretch}{1.05}

\begin{tabular*}{\linewidth}{
@{\extracolsep{\fill}}
lcccccc
@{}
}
\toprule
\multirow{2}{*}{\textbf{Method}}
& \multicolumn{3}{c}{\textbf{InterHuman$^{\dagger}$ + Inter-X$^{\dagger}$}}
& \multicolumn{3}{c}{\textbf{Unseen Motions}} \\
\cmidrule(lr){2-4}
\cmidrule(lr){5-7}
&
${E_{\mathit{mpjpe}}}\downarrow$
&
${E_{\mathit{inter\text{-}geo}}}\downarrow$
&
Succ(\%)$\uparrow$
&
${E_{\mathit{mpjpe}}}\downarrow$
&
${E_{\mathit{inter\text{-}geo}}}\downarrow$
&
Succ(\%)$\uparrow$
\\
\midrule

\rowcolor{gray!15}
\multicolumn{7}{l}{\textit{Single-Agent Trackers}} \\

TWIST2
& 86.56 & 11.44 & 41.36
& 113.35 & 14.72 & 28.58 \\

BeyondMimic
& 64.08 & 8.91 & 69.27
& -- & -- & -- \\

SONIC
& 65.31 & 9.56 & 63.32
& 66.31 & 9.51 & 62.65 \\

\midrule

\rowcolor{gray!15}
\multicolumn{7}{l}{\textit{Peer-Aware Controllers}} \\

From-Scratch RL
& 59.93 & 8.54 & 68.49
& 68.71 & 10.83 & 63.78 \\

Specialist$\rightarrow$Unified MLP
& 46.23 & 7.29 & 80.68
& 63.52 & 9.76 & 68.20 \\

\textbf{OmniInteract}
& \textbf{41.67} & \textbf{6.42} & \textbf{84.32}
& \textbf{61.67} & \textbf{8.92} & \textbf{72.89} \\

\bottomrule
\end{tabular*}
\end{table}
\subsection{Effectiveness: Partner-Aware Residual Control (Q2)}
\label{q2}

\begin{figure}[t]
    \centering
    \includegraphics[width=\linewidth]{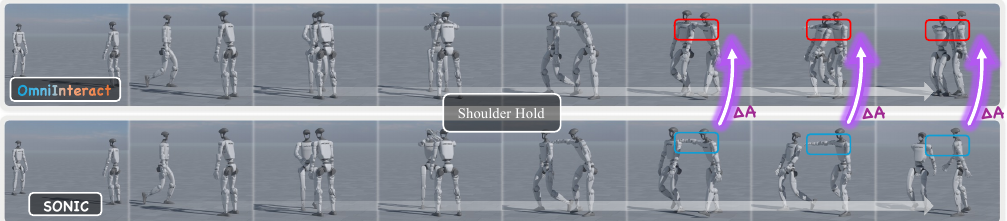}
    \caption{\textbf{Shoulder Hold in simulation (Issac Lab).}
Compared with SONIC (bottom), OmniInteract (top) improves
hand--shoulder alignment through partner-aware residual corrections.}
    \label{fig:sim_vis}
    \vspace{-15pt}
\end{figure}

\paragraph{Simulation Evaluation.}
To evaluate the effectiveness of partner-aware residual learning in leveraging single-humanoid motion priors for coordinated dual-humanoid interaction, we
compare OmniInteract with independent trackers and peer-aware controllers on
the selected InterHuman and Inter-X subset.
Table~\ref{tab:control} reveals a clear gap between individual motion tracking
and interactive coordination. Although single-agent trackers achieve reasonable
pose imitation, they fail to maintain interaction consistency under coupled
dynamics. In particular, SONIC attains comparable tracking accuracy but only
achieves $63.32\%$ success, indicating that independently following paired
references is insufficient to preserve partner-relative configurations. Figure~\ref{fig:sim_vis} illustrates this distinction:
SONIC reproduces the overall motion but exhibits hand--shoulder
misalignment, whereas OmniInteract maintains closer contact alignment. Moreover, From-scratch RL reaches only $68.49\%$ success, as it must jointly
discover individual tracking, balance, and interaction coordination. Building interaction capability on top of the frozen SONIC prior improves this limitation. In contrast, Specialist$\rightarrow$Unified MLP achieves $80.68\%$ success,
showing the benefit of learning interaction-specific corrections in a residual
space. OmniInteract further improves success to $84.32\%$ while achieving the
lowest interaction-geometry error ($6.42\%$) and $E_{mpjpe}$ ($41.67$\,mm).
These results demonstrate that preserving pretrained motor priors while
learning partner-dependent residual 
corrections provides a more effective
route to scalable dual-humanoid coordination.

\subsection{Scalability: Unified Control Cross Diverse Interactions (Q3) }
\label{q3}

\par\addvspace{\intextsep}
\noindent
\begin{minipage}{\linewidth}
\centering

\begin{minipage}[t]{0.5\linewidth}
\vspace{0pt}
\centering
\includegraphics[width=\linewidth]{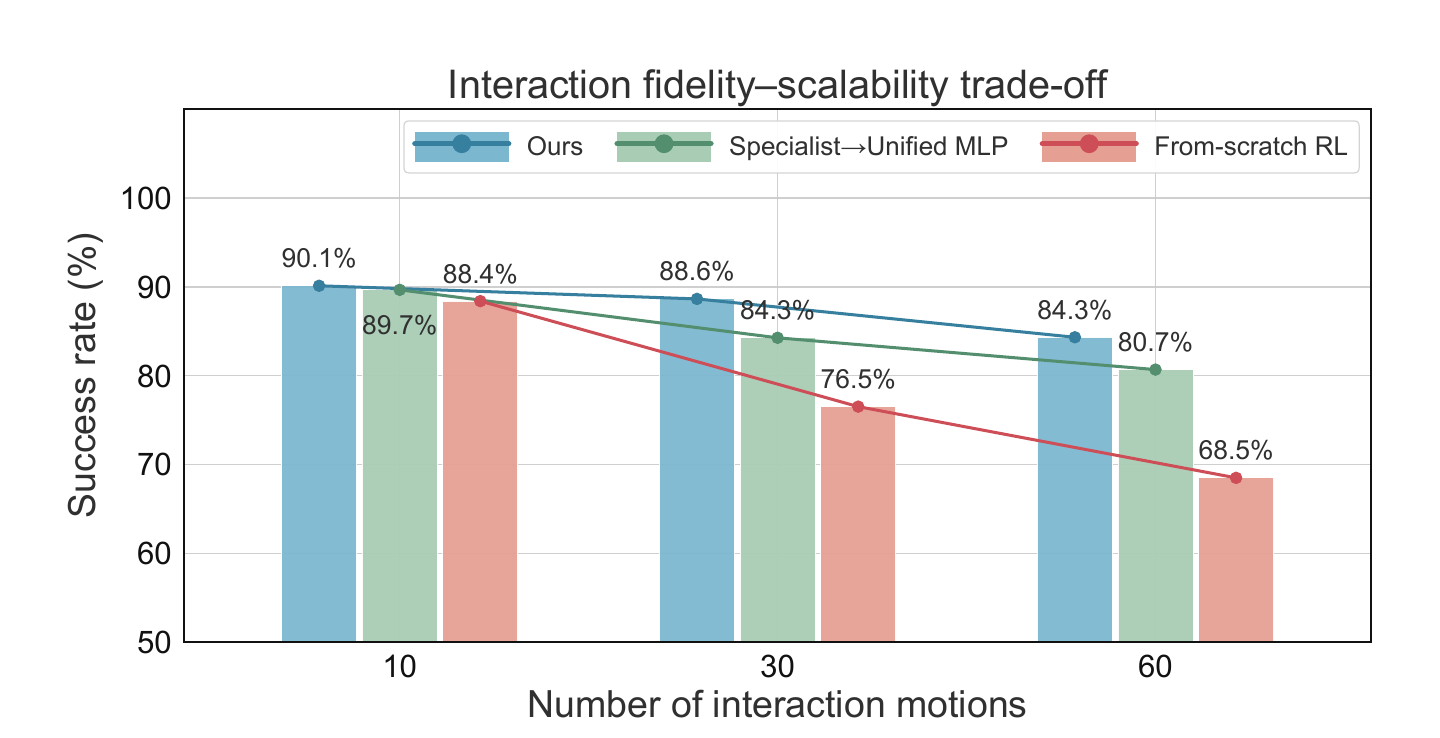}
\captionof{figure}{Fidelity--scalability trade-off.}
\label{fig:fidelity_scalability}
\end{minipage}
\hfill
\begin{minipage}[t]{0.45\linewidth}
\vspace{0pt}
\centering
\captionof{table}{Real-world evaluation on two Unitree G1 robots across three interaction tasks.}
\label{tab:real_world_g1}

{\scriptsize
\setlength{\tabcolsep}{2pt}
\renewcommand{\arraystretch}{1.10}

\begin{tabular*}{\linewidth}{
@{\extracolsep{\fill}}
>{\raggedright\arraybackslash}p{0.25\linewidth}|
>{\centering\arraybackslash}p{0.18\linewidth}
>{\centering\arraybackslash}p{0.23\linewidth}
>{\centering\arraybackslash}p{0.24\linewidth}
@{}
}
\toprule
\textbf{Task}
& \textbf{Attempts}
& \textbf{Method}
& \textbf{Succ. (\%)}$\uparrow$ \\
\midrule

\multirow{2}{*}{High-Five}
& \multirow{2}{*}{14}
& SONIC & 28.6 \\
& & \textbf{Ours} & \textbf{85.7} \\
\cmidrule(lr){1-4}

\multirow{2}{*}{Hug}
& \multirow{2}{*}{22}
& SONIC & 22.7 \\
& & \textbf{Ours} & \textbf{72.7} \\
\cmidrule(lr){1-4}

\multirow{2}{*}{Shoulder}
& \multirow{2}{*}{18}
& SONIC & 11.1 \\
& & \textbf{Ours} & \textbf{77.8} \\

\bottomrule
\end{tabular*}
}
\end{minipage}

\end{minipage}

To evaluate the generalization and scalability of unified interaction
control, we first test on unseen motions (randomly sampled from retargeted InterHuman \& Inter-X). As summarized in Table~\ref{tab:control}, OmniInteract outperforms
all applicable baselines across the all metrics, achieving
$72.89\%$ success and demonstrating that its coordination capabilities
transfer beyond the training references. To characterize the fidelity–scalability trade-off in multi-interaction control, we further scale the interaction diversity by training on an expanding set of motions drawn from our control benchmark. As shown in
Fig.~\ref{fig:fidelity_scalability}, the methods perform similarly at
small scale, but From-scratch RL deteriorates more rapidly as motion
diversity increases. At 60 motions, OmniInteract retains $84.3\%$ success,
compared with $80.7\%$ for Specialist$\rightarrow$Unified MLP and $68.5\%$
for From-scratch RL. These results support flow-based specialist
consolidation as an effective way to accommodate diverse interactions
while limiting the loss of execution fidelity.


\begin{wrapfigure}[20]{r}{0.5\columnwidth}
    \centering
    \includegraphics[width=\linewidth]{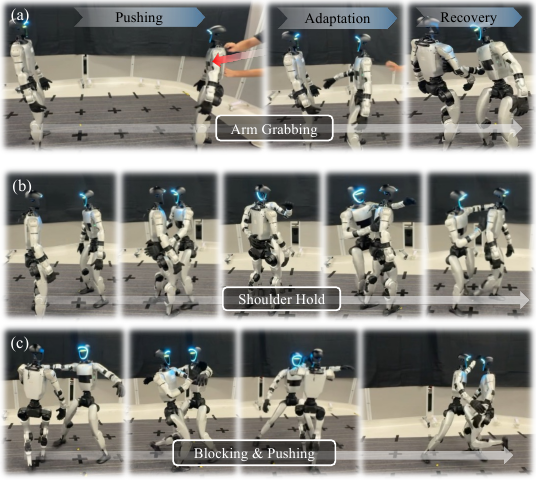}
    \caption{Real-world deployment of contact interaction tasks and robustness to disturbances.}
    \label{fig:real_world_g1}
\end{wrapfigure}
\subsection{Robustness: Real-World Dual-Humanoid Interaction (Q4)} 
\label{q4}
We deploy OmniInteract on two Unitree G1 humanoids and evaluate its real-world
performance on contact-rich interactions and under external force
perturbations. Table~\ref{tab:real_world_g1} reports success rates based on
contact establishment at specific keyframes across
three tasks: High-Five, Hug, and Shoulder Hold. OmniInteract consistently outperforms the single-humanoid tracker baseline SONIC by
over 50\%, with the largest margin on Shoulder Hold
(77.8\% vs.\ 11.1\%). SONIC is particularly vulnerable to accumulated
relative drift in this long-horizon interaction, while OmniInteract
continuously corrects partner-dependent misalignment through its residual corrections policy. These results demonstrate that partner-aware corrections transfer
effectively to physical robots, while independent tracking often fails to
establish the intended contact positions.

\paragraph{Robustness in Realistic Deployment.}
To evaluate robustness under external disturbances, we apply pushes during
interaction execution. As shown in Fig.~\ref{fig:real_world_g1}(a), the
humanoids adapt their poses and relative configuration to recover the intended
arm-grabbing interaction after perturbation. The Shoulder Hold and
Blocking \& Pushing examples in
Fig.~\ref{fig:real_world_g1}(b--c) further demonstrate stable coordination
under sustained physical coupling. These results highlight the closed-loop
correction capability of the partner-aware residual policy, which compensates
for interaction-induced deviations and restores task-relevant contacts online.

\section{Conclusion}

In this paper, we present \textbf{OmniInteract} to address the challenges
of preserving interaction geometry across embodiments and scaling
coordinated control across diverse dual-humanoid interactions.
We introduce SCAR, which jointly retargets both humanoids through contact-interface and near-field context modeling, producing
synchronized references that preserve fine-grained surface relationships.
We further propose a scalable flow-based interaction policy that factorizes control into a frozen motor prior and partner-dependent residuals and consolidates expert corrections into a unified generative controller through flow matching, while retaining pretrained motor capabilities.
Extensive experiments demonstrate improved retargeting fidelity,
higher coordination success, and better performance retention as
interaction diversity grows, with physical deployment validating
contact establishment and recovery from external perturbations.
By connecting surface interaction modeling with scalable residual
control, OmniInteract provides an effective route from human
demonstrations to coordinated physical interactions between humanoids.

\bibliography{iclr2027_conference}
\bibliographystyle{iclr2027_conference}

\clearpage
\appendix

\section*{Appendix}

\paragraph{Overview.}
This appendix provides implementation details and additional experimental
results for \textbf{OmniInteract}. It is organized into four sections (A--D):

\begin{itemize}[leftmargin=*]

\item \textbf{Section~\ref{app:scar}: Details of Surface Contact-Aware Retargeting.}
We describe motion standardization, surface correspondence, joint optimization
objectives and constraints, and solver settings, and explain the complementary
roles of contact-interface and near-field context representations.

\item \textbf{Section~\ref{app:sipl}: Details of Scalable Flow-Based Interaction Policy.}
We detail residual interaction expert learning and flow-based policy
consolidation, including observation spaces, network architectures, rewards,
training settings, and real-world deployment.

\item \textbf{Section~\ref{app:exp}: Experimental Details.}
We introduce the interaction motion benchmark and unseen-motion evaluation
set, describe baseline implementations, and specify the evaluation protocol
and metrics for retargeting and interaction control.

\item \textbf{Section~\ref{app:additional}: Additional Experiments.}
We present SCAR component ablations and qualitative retargeting comparisons,
followed by residual-scale sensitivity analysis and qualitative interaction
control results in simulation and the real world.

\end{itemize}
\section{Details of Surface Contact-Aware Retargeting}
\label{app:scar}

This section provides the implementation details of surface contact-aware
retargeting (SCAR), following the formulation and notation in the main text.
For humanoid $a\in\{1,2\}$, $\mathit h_{t,i}^{a}$ denotes a source surface
point and $\mathit x_i^{a}(q_t^{a})$ its corresponding robot surface point.
The paired configuration is $q_t=(q_t^1,q_t^2)$, and $\bar a=3-a$ denotes
the opposite humanoid. We first describe motion standardization and surface
binding, then detail the objectives and constraints used by the joint
optimization, and finally explain the complementary roles of the two
interaction representations.

\subsection{Motion Standardization and Surface Correspondence}
\label{app:scar_source}

\noindent\textbf{Standardization of motion sources.}
The InterHuman and Inter-X datasets provide synchronized, SMPL-X-compatible
two-person motions at 30\,Hz. Both actors are expressed in a common $z$-up
world frame, preserving their relative translations and headings. To adapt
the source pair to the target embodiment, we apply the same scale to both
actors:
\begin{equation}
 s=\frac{H_{\mathit{robot}}}{(H^1+H^2)/2},
 \label{eq:scar_app_scale}
\end{equation}
where $H^a$ is the canonical mesh height of source actor $a$ and
$H_{\mathit{robot}}$ is the robot height used by the correspondence pipeline.
The source points $\mathit h_{t,i}^{a}$ used throughout SCAR refer to the
surface samples \emph{after} this common-world standardization and scaling.
We do not independently recenter or rescale the two actors, since such
operations would change their demonstrated inter-body relationships.
The current source decoding configuration neutralizes finger articulation;
the objectives operate on the resulting body and hand surfaces.

\medskip
\noindent\textbf{Canonical correspondence and surface binding.}
We use the Dense Point Cloud Correspondence \citep{cao2026unified} adopted in the main text, together
with UMR's surface-binding implementation, to associate source and robot
surface samples in the canonical pose. A source sample is bound to a template
triangle with fixed barycentric coordinates and follows that triangle through
the motion. Its corresponding robot sample is projected onto robot geometry
and bound in the geometry's local frame. Consequently,
\begin{equation}
 \mathit x_i^a(q_t^a)
 =\mathit R_{b(i)}^a(q_t^a)\bar{\mathit x}_i^a
  +\mathit t_{b(i)}^a(q_t^a),
 \label{eq:scar_app_binding}
\end{equation}
where $b(i)$ identifies the rigid geometry carrying the sample, and
$\bar{\mathit x}_i^a$ is its fixed local position. The correspondence indices
remain unchanged over time; only the posed surface locations change.
This allows interaction relationships detected on the human surfaces to be
evaluated at the corresponding robot surfaces through forward kinematics.

\medskip
\noindent\textbf{Surface positions and orientations.}
The correspondence caches used in our evaluation contain 4,096 surface slots
per actor, where a slot is an indexed human--robot surface association.
For individual motion tracking, we select a fixed subset
$\mathcal I_{\mathit{pose}}^a$ with at most 12 slots per nonempty body region,
sampled without replacement. These indices are reused throughout the motion.
Contact and context construction instead query the complete surface pools
before applying their own budgets. Thus, the pose-sampling limit does not
restrict the number or location of candidate interaction points.
Surface orientations are represented by normals: the target robot normals
$\widehat{\mathit n}_{t,i}^a$ are obtained by calibrating canonical robot
normals and transporting them with the source surface deformation, following
UMR, rather than directly copying absolute human normals.

\subsection{Optimization Details and Hyperparameters}
\label{app:scar_optimization}

\noindent\textbf{Joint retargeting formulation.}
Following the main text, SCAR approximately solves the frame-wise problem
\begin{equation}
 q_t^\star
 =\underset{q_t\in\mathcal C_{\mathit{phy}}}{\operatorname{arg\,min}}
 \;\mathcal E_{\mathit{pose}}(q_t)
 +\lambda_{\mathit c}\mathcal E_{\mathit{contact}}(q_t)
 +\lambda_{\mathit n}\mathcal E_{\mathit{context}}(q_t)
 +\mathcal R_t(q_t).
 \label{eq:scar_app_objective}
\end{equation}
Both robots' root poses and joint configurations are optimization variables.
The contact and context terms couple the two configurations, while
$\mathcal C_{\mathit{phy}}$ accounts for joint, velocity, floor, and collision
constraints. Below, we specify each term and the numerical realization of
these constraints.

\medskip
\noindent\textbf{Individual motion preservation.}
The pose objective matches the selected surface positions and normal
orientations:
\begin{equation}
 \begin{aligned}
 \mathcal E_{\mathit{pose}}(q_t)
 =\sum_{a\in\{1,2\}}\sum_{i\in\mathcal I_{\mathit{pose}}^a}
 \Big(&w_i^{a,\mathit{pos}}
       \|\mathit x_i^a(q_t^a)-\mathit h_{t,i}^a\|_2^2\\
     +&w_i^{a,\mathit{nor}}
       \|\mathit n_i^a(q_t^a)-\widehat{\mathit n}_{t,i}^a\|_2^2\Big).
 \end{aligned}
 \label{eq:scar_app_pose}
\end{equation}
Here $w_i^{a,\mathit{pos}}$ and $w_i^{a,\mathit{nor}}$ are region-dependent
weights inherited from UMR. They are distinct from the neighborhood weights
$\omega_{t,ij}$ in the interaction coordinates.

\begin{figure}[t]
    \centering
    \includegraphics[width=0.65\linewidth]{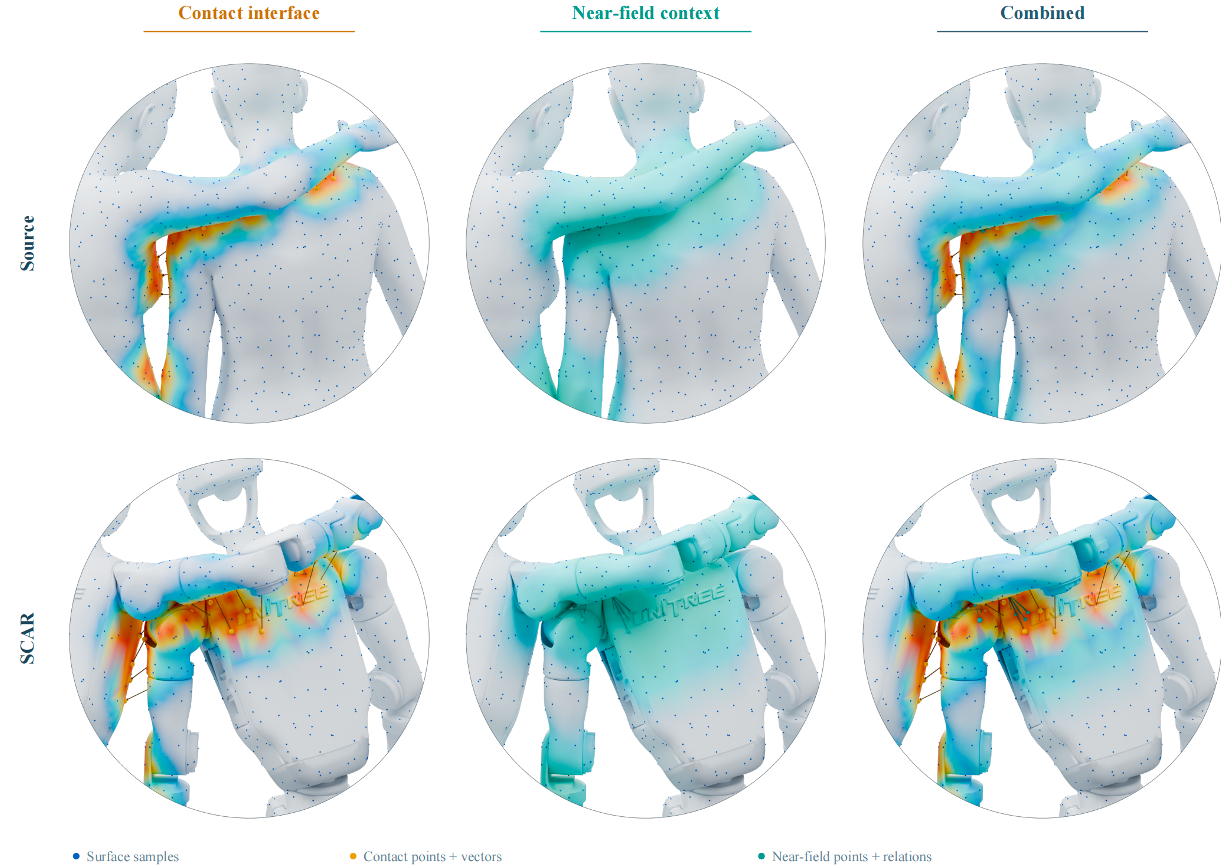}
\caption{
Visualization of contact interface and near-field context.
}
    \label{fig:app1}
\end{figure}

\noindent\textbf{Contact-Interface Objective.}
We construct $\mathcal C_t$ from the union of directed nearest-neighbor
candidates in both directions, $1\!\rightarrow\!2$ and
$2\!\rightarrow\!1$, together with the retained pairs from the previous
frame. Recall the demonstrated contact offset
$\mathit d_{t,ij}^{h}=\mathit h_{t,i}^{1}-\mathit h_{t,j}^{2}$, and let
$\ell_{t,ij}=\|\mathit d_{t,ij}^{h}\|_2$ be its length.
A new candidate is eligible if $\ell_{t,ij}\leq\tau_{\mathit{in}}$; a
previously retained pair remains eligible while
$\ell_{t,ij}\leq\tau_{\mathit{out}}$, with
$\tau_{\mathit{in}}<\tau_{\mathit{out}}$. This hysteresis reduces contact
switching near the threshold. Eligible pairs are grouped by their two body
regions, ordered by distance within each group, and selected in round-robin
order across groups until the contact budget is reached.

The normalized source weights in the main-text contact objective are
\begin{equation}
 \widetilde\alpha_{t,ij}
 =\operatorname{clip}\!\left(
 \frac{\tau_{\mathit{out}}-\ell_{t,ij}}
      {\tau_{\mathit{out}}-\tau_{\mathit{in}}},\,0.1,\,1\right),
 \qquad
 \alpha_{t,ij}
 =\frac{\widetilde\alpha_{t,ij}}
        {\sum_{(u,v)\in\mathcal C_t}\widetilde\alpha_{t,uv}}.
 \label{eq:scar_app_alpha}
\end{equation}
Using $\mathit d_{t,ij}^{r}(q_t)=
\mathit x_i^1(q_t^1)-\mathit x_j^2(q_t^2)$, we minimize
\begin{equation}
 \mathcal E_{\mathit{contact}}(q_t)
 =\sum_{(i,j)\in\mathcal C_t}\alpha_{t,ij}
 \|\mathit d_{t,ij}^{r}(q_t)-\mathit d_{t,ij}^{h}\|_2^2.
 \label{eq:scar_app_contact}
\end{equation}
Each contact pair contributes an explicit relative-vector residual,
penalizing both separation and tangential misalignment without averaging
errors across the interface. The target is the demonstrated offset, not a
forced zero vector, so small source clearances are retained. Both endpoints
are movable: if $\mathit J_{t,i}^a$ denotes the surface-point Jacobian with
respect to robot $a$'s tangent coordinates, the contact residual has Jacobian
$[\mathit J_{t,i}^1,-\mathit J_{t,j}^2]$. Thus, contact preservation can be
achieved through coordinated changes to both robots. It remains a soft
geometric objective, not a contact-force or exact zero-gap constraint.

\medskip
\noindent\textbf{Near-Field Context Objective.}
For each source point not retained as a contact endpoint, define its distance
to the opposite surface as
$\ell_{t,i}^a=\min_j\|\mathit h_{t,i}^a-\mathit h_{t,j}^{\bar a}\|_2$.
Points within radius $r_{\mathit n}$ are eligible context centers. We jointly
rank candidates from both actors by $\ell_{t,i}^a$ and retain them under a
shared budget. $\mathcal S_t^a$ denotes the retained centers belonging to
actor $a$; hence $|\mathcal S_t^1|+|\mathcal S_t^2|\leq B_{\mathit n}$.

For each center $i\in\mathcal S_t^a$, we find the nearest point on actor
$\bar a$ and its one-ring neighbors in the cached correspondence topology.
Among these points, the closest at most $K$ lying within $r_{\mathit n}$ of
the center form $\mathcal N_t(i)$. Actor identity is implicit in this notation:
all $j\in\mathcal N_t(i)$ belong to $\bar a$. The normalized source-geometric
weights are
\begin{equation}
 \omega_{t,ij}
 =\frac{\exp\!\left(-\|\mathit h_{t,i}^a-
                     \mathit h_{t,j}^{\bar a}\|_2^2/(2\sigma_{\mathit n}^2)\right)}
 {\sum_{u\in\mathcal N_t(i)}
  \exp\!\left(-\|\mathit h_{t,i}^a-
                     \mathit h_{t,u}^{\bar a}\|_2^2/(2\sigma_{\mathit n}^2)\right)}.
 \label{eq:scar_app_omega}
\end{equation}
As in the main text, the interaction coordinate of a human or robot point
$\mathit p_i$ is
\begin{equation}
 \mathcal L_t(\mathit p_i)
 =\mathit p_i-\sum_{j\in\mathcal N_t(i)}\omega_{t,ij}\mathit p_j,
 \qquad \sum_{j\in\mathcal N_t(i)}\omega_{t,ij}=1.
 \label{eq:scar_app_laplacian}
\end{equation}
This coordinate measures the displacement from a weighted neighborhood
centroid on the opposite body to the center point. For robot coordinates,
the center is evaluated at $q_t^a$ and all neighbors at $q_t^{\bar a}$;
therefore $\mathcal L_t(\mathit x_i^a(q_t^a))$ depends on the paired
configuration despite being indexed by its center.
The neighborhoods and weights are constructed from the source geometry and
held fixed throughout the current frame's optimization. Contact endpoints
are excluded only as context \emph{centers}; they can still serve as neighbors.

The center-proximity weights are normalized across both humanoids:
\begin{equation}
 \widetilde\beta_{t,i}^a
 =\exp\!\left(-\frac{(\ell_{t,i}^a)^2}{2r_{\mathit n}^2}\right),
 \qquad
 \beta_{t,i}^a
 =\frac{\widetilde\beta_{t,i}^a}
 {\sum_{b\in\{1,2\}}\sum_{u\in\mathcal S_t^b}\widetilde\beta_{t,u}^b}.
 \label{eq:scar_app_beta}
\end{equation}
The context objective is then
\begin{equation}
 \mathcal E_{\mathit{context}}(q_t)
 =\sum_{a\in\{1,2\}}\sum_{i\in\mathcal S_t^a}\beta_{t,i}^a
 \left\|
 \mathcal L_t\!\left(\mathit x_i^a(q_t^a)\right)
 -\mathcal L_t\!\left(\mathit h_{t,i}^a\right)
 \right\|_2^2.
 \label{eq:scar_app_context}
\end{equation}
This term preserves the surrounding surface arrangement through local
cross-body coordinates rather than requiring every nearby pair to match
independently. If $\mathcal C_t$ or both context-center sets are empty, the
corresponding interaction term is set to zero without evaluating its weight
normalization. Fig.~\ref{fig:app1} shows the visualization of contact interface and near-field context.

\medskip
\noindent\textbf{Temporal and stance-foot regularization.}
We instantiate the main-text regularizer as
\begin{equation}
 \mathcal R_t(q_t)
 =\lambda_{\mathit{temp}}
  \left\|\frac{q_t\ominus q_{t-1}^\star}{\Delta t}\right\|_2^2
 +\lambda_{\mathit{foot}}\mathcal E_{\mathit{foot}}(q_t),
 \label{eq:scar_app_regularization}
\end{equation}
where $\ominus$ denotes a tangent-space configuration difference. The first
term penalizes generalized velocity magnitude and is omitted at the first
frame. A source foot is marked as supporting when its foot joint lies less
than 8\,cm above that actor's lowest foot-joint height over the clip and moves
slower than 0.25\,m/s. At each support onset, up to eight robot foot-surface
points are recorded from the frame initialization as fixed anchors
$\bar{\mathit x}_{f,i}^{\mathit{foot}}$. For the active supporting feet
$f\in\mathcal B_t$, with nonempty anchor-index sets
$\mathcal I_f^{\mathit{foot}}$, we use
\begin{equation}
 \mathcal E_{\mathit{foot}}(q_t)
 =\sum_{f\in\mathcal B_t}\frac{1}{|\mathcal I_f^{\mathit{foot}}|}
  \sum_{i\in\mathcal I_f^{\mathit{foot}}}
  \|\mathit x_{f,i}(q_t)-\bar{\mathit x}_{f,i}^{\mathit{foot}}\|_2^2.
 \label{eq:scar_app_foot}
\end{equation}
This softly discourages three-dimensional stance-foot drift; it does not
impose a hard no-slip constraint.

\medskip
\noindent\textbf{Constraints and SQP subproblems.}
We implement SQP using damped Gauss--Newton quadratic subproblems. Let
$\mathit e_t(q_t)$ stack the weighted residuals of
~\redeqref{eq:scar_app_objective}, and let $\mathit J_t$ be their
tangent-space linearization. At each inner iteration, we solve
\begin{equation}
 \min_{\Delta q_t,\,\xi\geq0}
 \frac12\|\mathit e_t(q_t)+\mathit J_t\Delta q_t\|_2^2
 +\frac\mu2\|\Delta q_t\|_2^2
 +\frac{\lambda_{\mathit{slack}}}{2}\|\xi\|_2^2,
 \label{eq:scar_app_qp}
\end{equation}
subject to the following conditions, which numerically realize
$\mathcal C_{\mathit{phy}}$.
\begin{itemize}
 \item \textbf{Joint and velocity limits.}
 The actuated joint coordinates are bounded by the robot model's joint
 ranges. Componentwise tangent bounds additionally enforce
 $-v_{\max}\Delta t\leq(q_t\ominus q_{t-1}^\star)+\Delta q_t
 \leq v_{\max}\Delta t$ in the local approximation, using separate limits
 for root translation, root rotation, and actuated joints. Velocity bounds
 are omitted at the first frame.
 \item \textbf{Floor and collision constraints.}
 For each active geometric row, let $\phi_\ell(q_t)$ denote a sampled
 surface height or a signed collision-geometry distance, with margin
 $m_\ell$. We impose
 $\phi_\ell(q_t)+J_{\phi_\ell}\Delta q_t\geq m_\ell-\xi_\ell$.
 The nonnegative slack $\xi_\ell$ carries a large penalty, allowing recovery
 from initially violated configurations. Collision checks include self- and
 inter-robot pairs, excluding same-actor body pairs within three
 kinematic-tree edges.
 \item \textbf{Trust region and tolerance envelope.}
 Each update satisfies $\|\Delta q_t\|_\infty\leq\delta$.
 Protected geometric rows also obey the hard, slack-free bound
 $\phi_\ell(q_t)+J_{\phi_\ell}\Delta q_t
 \geq m_\ell-\varepsilon_{\mathit{env}}$, where
 $\varepsilon_{\mathit{env}}=0.9\varepsilon$ and $\varepsilon$ is the
 acceptance tolerance. All active geometric rows are protected when the
 current iterate passes the feasibility check; otherwise, only rows already
 inside this tighter envelope are protected.
\end{itemize}

\noindent\textbf{Initialization and nonlinear acceptance.}
The first frame uses source-derived root positions and headings with a fixed
bent-arm initialization. Subsequent frames are warm-started from the previous
solution $q_{t-1}^\star$. After each QP solve, we evaluate
$q_t^+=q_t\oplus\zeta\Delta q_t$ with
$\zeta\in\{1,1/2,\ldots,1/64\}$, where $\oplus$ integrates the floating-base
rotations on their configuration manifold. When the current iterate is
feasible, a candidate must stay within the acceptance tolerance and not
increase the tracking objective in Eq.~\redeqref{eq:scar_app_objective}. When it
is infeasible, the rule prioritizes decreasing violation, permitting objective
descent when violation is unchanged within numerical tolerance. If no step is
accepted, the current iterate is retained and the frame solve terminates.
Final validity flags are saved alongside the trajectories; invalid frames are
not silently removed.

\medskip
\noindent\textbf{Implementation and hyperparameters.}
We use MuJoCo \citep{todorov2012mujoco} for forward kinematics, surface Jacobians, and collision
queries, and UMR's \citep{cao2026unified} Clarabel interface to solve the quadratic subproblems.
The paired 29-DoF G1 configuration has 72 stored coordinates, including root
quaternions, and a 70-dimensional tangent update. The settings in
Tables~\ref{tab:scar_app_parameters} and~\ref{tab:scar_app_pose_weights} are
shared by the evaluated InterHuman and Inter-X configurations.
Contact weights $\alpha_{t,ij}$ sum to one over $\mathcal C_t$;
neighbor weights $\omega_{t,ij}$ sum to one within each neighborhood;
center weights $\beta_{t,i}^a$ sum to one across the retained centers of both
actors. The pose objective instead sums its region-weighted residuals.

The geometric constraints use floor and collision margins of $-2$ and
$-3$\,mm, respectively, and a 5\,mm acceptance tolerance relative to these
margins. Their effective nonlinear bounds are therefore $z\geq-7$\,mm and
$d_{\mathit{signed}}\geq-8$\,mm; the hard envelope leaves a further 0.5\,mm
buffer. These are numerical tolerances, not exact non-penetration guarantees.
The current implementation compares the numerical maximum of geometric
margin violations and componentwise speed excess against the same tolerance.
Its validity flag is consequently an implementation-level diagnostic rather
than a physical-executability certificate. SCAR generates coupled kinematic
references; sustaining their execution under dynamics is the role of the
interaction-policy stage of OmniInteract.

\begin{table}[t]
\centering
\small
\caption{SCAR interaction and optimization settings. Context centers share
one budget across the two humanoids.}
\label{tab:scar_app_parameters}
\begin{tabular}{@{}ll@{}}
\toprule
Parameter & Value \\
\midrule
Contact thresholds $\tau_{\mathit{in}}/\tau_{\mathit{out}}$ & $0.02/0.03$\,m \\
Contact-pair budget & $128$ per frame \\
Context-center budget $B_{\mathit n}$ & $128$ total across both actors \\
Context radius $r_{\mathit n}$ / Gaussian width $\sigma_{\mathit n}$ & $0.15/0.06$\,m \\
Maximum neighbors $K$ & $6$ per center \\
Pose slots per region / sampling seeds & $12$ / $0,1$ for actors $1,2$ \\
$\lambda_{\mathit c}/\lambda_{\mathit n}$ & $2000/100$ \\
$\lambda_{\mathit{temp}}/\lambda_{\mathit{foot}}$ & $0.02/50$ \\
Damping $\mu$ / slack penalty $\lambda_{\mathit{slack}}$ & $0.01/10^4$ \\
Step bound $\delta$ & $0.12$\,m or $0.12$\,rad by coordinate \\
Joint / root angular speed limits & $8/4$\,rad/s \\
Root linear speed limit & $2$\,m/s \\
Floor / collision activation distance & $0.04/0.04$\,m \\
Maximum active floor samples & $128$ per actor \\
First-frame / later-frame SQP iterations & $30/8$ \\
Backtracking scales $\zeta$ & $2^{-k}$, $k=0,\ldots,6$ \\
\bottomrule
\end{tabular}
\end{table}

\begin{table}[t]
\centering
\small
\caption{Region-dependent pose weights inherited from UMR. Left and right
regions share weights. \texttt{Arm} and \texttt{UpperArm} denote distinct
regions in the 19-region configuration.}
\label{tab:scar_app_pose_weights}
\begin{tabular}{@{}lrr@{}}
\toprule
Region & Position $w^{\mathit{pos}}$ & Normal $w^{\mathit{nor}}$ \\
\midrule
Torso & 20 & 5 \\
Hips & 20 & 0 \\
Head & 1 & 1 \\
Hand, Foot & 10 & 1 \\
Arm, ForeArm, UpLeg, Leg & 10 & 5 \\
Shoulder & 0 & 0.5 \\
UpperArm & 0 & 0.1 \\
\bottomrule
\end{tabular}
\end{table}

\subsection{Surface Interaction Graph and Complementary Roles}

SCAR's interaction representation separates \emph{where the surfaces should
interact} from \emph{how the surrounding surfaces should be arranged}.
Contact-interface vectors connect individual pairs in $\mathcal C_t$ and
can be displayed in orange--gold. They directly visualize
$\mathit d_{t,ij}^{h}$ and $\mathit d_{t,ij}^{r}(q_t)$ on the source and
robot surfaces. Near-field relations can be displayed in teal: neighborhood
spokes identify the points in $\mathcal N_t(i)$, and the arrow from their
weighted centroid to the center depicts $\mathcal L_t(\mathit p_i)$.
The spokes contribute to one weighted interaction coordinate; they are not
separate pairwise objectives. Surface heatmaps highlight proximity to the
displayed graph endpoints and should not be interpreted as forces or
optimization residuals.

In an embrace, for example, contact pairs specify the palm--back and
arm--torso interfaces, while context centers describe the arrangement of
nearby limbs and torso regions. Matching the contact pairs alone leaves this
surrounding arrangement underconstrained. Conversely, a neighborhood's
weighted coordinate may remain similar even when individual contact offsets
deviate. Combining $\mathcal E_{\mathit{contact}}$ and
$\mathcal E_{\mathit{context}}$ addresses these two complementary requirements
within the same paired configuration, rather than independently retargeting
two motions and subsequently repairing their interaction. The resulting
trajectories provide synchronized robot-space references for OmniInteract's
interaction-policy training. Algorithm~\ref{alg:scar_app} summarizes the
retargeting procedure.

\begin{algorithm}[t]
\caption{Surface Contact-Aware Retargeting (SCAR)}
\label{alg:scar_app}
\small
\begin{algorithmic}[1]
\Require Synchronized source motions, robot models, canonical surface correspondences
\Ensure Paired configurations $\{q_t^\star\}_{t=1}^T$ and validity diagnostics
\State \textbf{Phase 1: Shared surface preparation}
\State Standardize both actors; apply the shared scale $s$
\State Construct $\mathit h_{t,i}^a$, bindings for $\mathit x_i^a(q_t^a)$, and normal targets
\State Fix $\mathcal I_{\mathit{pose}}^a$; identify source foot-support intervals
\State Initialize $q_{\mathit{init}}$ and $\mathcal C_0\gets\varnothing$
\State \textbf{Phase 2: Joint frame-wise SQP}
\For{$t=1,\ldots,T$}
 \State Initialize $q_t\gets q_{t-1}^\star$; use $q_{\mathit{init}}$ if $t=1$
 \State Construct $\mathcal C_t$ with bidirectional search, hysteresis, and region balancing
 \State Compute contact weights $\alpha_{t,ij}$
 \State Jointly retain $\mathcal S_t^1,\mathcal S_t^2$ with $|\mathcal S_t^1|+|\mathcal S_t^2|\leq B_{\mathit n}$
 \State Build $\mathcal N_t(i)$, $\omega_{t,ij}$, $\beta_{t,i}^a$, and source coordinates $\mathcal L_t(\mathit h_{t,i}^a)$
 \State Update stance anchors; keep the source-defined interaction targets fixed
 \For{each SQP iteration}
  \State Evaluate the paired residuals, Jacobians, and active constraints
  \State Solve ~\redeqref{eq:scar_app_qp} with bounds and the hard tolerance envelope
  \If{the solve fails or $\|\Delta q_t\|_\infty<10^{-6}$}
   \State Terminate the current frame solve
  \EndIf
  \State Backtrack over $q_t\oplus\zeta\Delta q_t$ using nonlinear acceptance checks
  \If{a candidate is accepted}
   \State Update $q_t$ to the accepted candidate
  \Else
   \State Retain $q_t$ and terminate the current frame solve
  \EndIf
 \EndFor
 \State Store $q_t^\star\gets q_t$ and its validity diagnostics
\EndFor
\State \Return Paired robot trajectories and diagnostics
\end{algorithmic}
\end{algorithm}

\section{Details of Scalable Flow-Based Interaction Policy}
\label{app:sipl}

\subsection{Residual Interaction Expert Learning}
\label{app:sipl_experts}

\paragraph{Partner-aware residual control.}
SCAR provides synchronized robot-space kinematic references for the two
humanoids. We learn to execute these references under simulated dynamics
by augmenting the frozen SONIC tracker
$\pi_{\mathit{base}}$~\citep{luo2026sonic} with a task-specific residual
expert $\pi^k_{\textit{res}}$.
For each interaction task $k$, the expert is shared by both humanoids
and evaluated using their respective observations.
Following \redeqref{eq:interaction_residual}, the final action
$\mathcal A^a$ combines the frozen tracking action
$\mathcal A^a_{\mathit{base}}$ with the partner-dependent correction
$\Delta\mathcal A^{\bar a\rightarrow a}$, where
$a\in\{1,2\}$ and $\bar a=3-a$.

The base and residual actions use consistent joint ordering and
normalization.
We set $\alpha_{\mathit{res}}=0.3$ in normalized action coordinates
and zero-initialize the residual output layer.
Consequently, the initial deterministic controller reproduces the
frozen tracker, while subsequent optimization learns bounded
interaction-specific corrections.
Joint targets are updated at $50$\,Hz and tracked by implicit PD
actuators in a $200$\,Hz physics simulation.
Below, $n$ denotes a control step.

\paragraph{Observation space.}
The observation groups
$\mathcal O^a_{his}$, $\mathcal O^a_{ego}$, and
$\mathcal O^a_{fur}$ retain the definitions in the main text.
They are implemented as two temporal streams together with a
current-frame input to the action decoder.

\begin{itemize}
    \item \textbf{History,
    $\mathcal O^a_{his}\in\mathbb R^{20\times239}$.}
    The history covers control steps $n-19,\ldots,n$.
    Each entry contains ego tracking features ($64$), ego
    proprioception and the previous complete action ($93$),
    partner-state features ($64$), and reference and measured
    relative torso poses ($18$).
    Ego tracking comprises reference joint positions and velocities
    ($29$ each) and torso orientation error ($6$).
    Proprioception comprises projected gravity ($3$), base angular
    velocity ($3$), centered joint positions and velocities
    ($29$ each), and the previous action ($29$).

    \item \textbf{Ego-centric partner information,
    $\mathcal O^a_{ego}\in\mathbb R^{82}$.}
    This block contains the partner's reference and measured joint
    positions ($29$ each), torso orientation error ($6$), and
    reference and measured relative torso poses ($9$ each).
    Each relative pose consists of translation ($3$) and rotation
    ($6$), expressed in the corresponding ego torso frame.
    This block is included in the latest history entry rather
    than processed by an additional temporal encoder.

    \item \textbf{Future references,
    $\mathcal O^a_{fur}\in\mathbb R^{20\times93}$.}
    The next $20$ reference frames provide ego joint positions
    and velocities ($29$ each), future torso orientation relative
    to the current torso ($6$), and partner joint positions ($29$).
    Desired relative-pose information is supplied separately
    through $\mathcal O^a_{ego}$ and its history.
\end{itemize}


\paragraph{Network architecture and optimization.}
Two independent temporal 1D CNNs encode the normalized history and
future streams into $64$-dimensional embeddings.
A learned three-dimensional projection of the history embedding is
appended to that embedding, yielding $67$ dimensions; this projection
has no separate auxiliary supervision.
The residual decoder concatenates the history embedding ($67$),
future embedding ($64$), and latest history entry ($239$), including
$\mathcal O^a_{ego}$, into a $370$-dimensional input.

We train the shared expert with MAPPO under centralized training
and decentralized execution.
The centralized critic receives both humanoids' tracking states
and temporal inputs.
Each actor evaluation uses its own observations, communicated
partner information, and paired motion references.
During training, a diagonal Gaussian is centered on the concatenated
corrected actions of both humanoids, giving a $58$-dimensional joint
action distribution.
Thus, $\alpha_{\mathit{res}}$ bounds the deterministic residual
correction, not the sampled exploration noise.
PPO uses the joint action likelihood ratio and updates the residual
expert and critic while keeping SONIC frozen.
The core settings are summarized in
Table~\ref{tab:sipl_expert_settings}.

\begin{table}[t]
\centering
\caption{Network architecture and MAPPO hyperparameters for
residual interaction expert learning.}
\label{tab:sipl_expert_settings}
{\small
\setlength{\tabcolsep}{5pt}
\renewcommand{\arraystretch}{1.08}
\begin{tabular}{@{}ll@{}}
\toprule
\textbf{Hyperparameter} & \textbf{Value} \\
\midrule
\multicolumn{2}{c}{\textit{Architecture}} \\
History / future window length & $20$ / $20$ \\
Input projection, history / future
    & $239\rightarrow60$ / $93\rightarrow60$ \\
Temporal CNN channels & $60\rightarrow40\rightarrow20$ \\
CNN kernel sizes / strides & $(6,4)$ / $(2,2)$ \\
Temporal pooling / output projection
    & Average / $20\rightarrow64$ \\
Residual decoder input / output & $370$ / $29$ \\
Actor and critic hidden dimensions & $[512,256,128]$ \\
Hidden activation & RELU \\
Residual output initialization / scale & Zero / $0.3$ \\
\midrule
\multicolumn{2}{c}{\textit{MAPPO training}} \\
Rollout steps per environment & $24$ \\
Learning epochs / mini-batches & $5$ / $4$ \\
Optimizer / initial learning rate & Adam / $10^{-4}$ \\
Learning-rate schedule / target KL & Adaptive / $0.02$ \\
PPO clip parameter & $0.1$ \\
Entropy / value-loss coefficient & $0.001$ / $1.0$ \\
Discount factor $\gamma$ / GAE parameter $\lambda$
    & $0.99$ / $0.95$ \\
Clipped value loss & Enabled \\
Maximum gradient norm & $0.5$ \\
Initial action standard deviation & $0.1$ \\
\bottomrule
\end{tabular}
}
\end{table}

\paragraph{Interaction-aware rewards.}
We train the residual experts with a shared reward that balances
interaction geometry, contact timing, individual tracking, and control
regularization:
\begin{equation}
r_n
=
\Delta t_{\mathit{ctrl}}
\left(
r_{\mathit{inter}}
+\sum_{a=1}^{2}r_{\mathit{track}}^a
+r_{\mathit{reg}}
\right),
\qquad
\Delta t_{\mathit{ctrl}}=0.02\,\mathrm{s}.
\label{eq:sipl_expert_reward}
\end{equation}
Each component is the weighted sum of the corresponding block in
Table~\ref{tab:sipl_rewards}, including the negative regularization
weights. Tracking rewards are evaluated separately for each humanoid;
interaction rewards and penalties are evaluated once for the pair.
The action-rate penalty acts on the complete applied action
$\mathcal A_n^a$, rather than the residual alone.

Superscripts $\mathit{sim}$ and $\mathit{ref}$ denote simulated
quantities and targets from the synchronized retargeted robot
trajectories. We use $\mathbf p_{a,b}$ and $Q_{a,b}$ for body position
and unit quaternion, with the torso as the anchor $\mathit{anc}$.
The relative quantities are
$\Delta\mathbf p=\mathbf p_{1,\mathit{anc}}-\mathbf p_{2,\mathit{anc}}$
in world coordinates and
$\Delta Q=Q_{1,\mathit{anc}}^{-1}\otimes Q_{2,\mathit{anc}}$,
where $\otimes$ denotes quaternion multiplication and $d_Q$ measures
the relative rotation angle between the represented orientations.
A tilde denotes a reference aligned to the simulated torso's horizontal
position using the yaw component of the reference-to-simulated torso
rotation difference, while preserving reference torso height.
Velocity errors are evaluated in world coordinates. The quantities
$\mathbf v$ and $\boldsymbol\omega$ denote linear and angular
velocities, and the superscript $\mathit{xy}$ denotes horizontal components.

The tracked set $\mathcal B_{\mathit{track}}$ contains the pelvis,
torso, and bilateral hip-roll, knee, ankle-roll, shoulder-roll, elbow,
and wrist-yaw links.
The topology set $\mathcal P_{\mathit{topo}}$ contains the five
corresponding cross-humanoid pairs at the left wrist, right wrist,
torso, left ankle, and right ankle.
For each pair,
$\ell_{uv}=\|\mathbf p_{1,u}-\mathbf p_{2,v}\|_2$ and
$\rho_{uv}=1/\max(\ell_{uv}^{\mathit{ref}},0.1)$.

The set $\mathcal P_{\mathit{cont}}$ contains the two wrist pairs.
Their binary proximity labels $c_{uv}^{\mathit{sim}}$ and
$c_{uv}^{\mathit{ref}}$ use a distance threshold of $0.18$\,m.
For the two feet $\mathcal F$, reference support
$s_{a,f}^{\mathit{ref}}$ is determined by ankle height below $0.06$\,m,
and simulated support $s_{a,f}^{\mathit{sim}}$ by contact-force
magnitude above $1$\,N in any of the three stored sensor samples.
The contact reward jointly exponentiates the separately averaged
wrist-proximity and foot-support errors.

Finally, $q_{a,j}^{\mathit{act}}$ denotes an actuated joint position,
$q_{a,j}^{\mathit{min}}$ and $q_{a,j}^{\mathit{max}}$ are soft joint
limits, and $[x]_+=\max(x,0)$.
The set $\mathcal U$ contains monitored bodies other than wrist and
ankle links; $\mathbf F_{a,u,h}^{\mathit{sim}}$ denotes the measured
net contact force at history index $h$, and
$\boldsymbol\tau_a$ denotes joint torques.
This fixed contact penalty is a soft regularizer, not a
reference-conditioned exclusion rule.
These body-level execution rewards complement SCAR's kinematic
surface objectives without reusing its surface graph or contact
thresholds.

\begin{table}[t]
\centering
\caption{Reward terms and weights for residual interaction expert learning.}
\label{tab:sipl_rewards}
{\small
\setlength{\tabcolsep}{4pt}
\renewcommand{\arraystretch}{1.20}
\begin{tabularx}{\linewidth}{
@{}
>{\raggedright\arraybackslash}p{0.15\linewidth}
>{\raggedright\arraybackslash}p{0.195\linewidth}
>{\footnotesize\raggedright\arraybackslash}X
r
@{}
}
\toprule
\textbf{Category}
& \textbf{Reward term}
& {\small\textbf{Equation}}
& \textbf{Weight} \\
\midrule

\multirow[t]{4}{=}{\textbf{Interaction}}
& Relative anchor position
& $\exp\!\left(
  -\frac{\|\Delta\mathbf p^{\mathit{sim}}
  -\Delta\mathbf p^{\mathit{ref}}\|_2^2}{0.5^2}
  \right)$
& $1.5$ \\[3pt]

& Relative anchor orientation
& $\exp\!\left(
  -\frac{d_Q(\Delta Q^{\mathit{sim}},
  \Delta Q^{\mathit{ref}})^2}{0.75^2}
  \right)$
& $0.5$ \\[3pt]

& Interaction topology
& $\exp\!\left(
  -\frac{\sum_{(u,v)\in\mathcal P_{\mathit{topo}}}
  \rho_{uv}(\ell_{uv}^{\mathit{sim}}
  -\ell_{uv}^{\mathit{ref}})^2}
  {0.35^2\,|\mathcal P_{\mathit{topo}}|}
  \right)$
& $1.5$ \\[3pt]

& Contact/support timing
& $\exp\!\left(\begin{aligned}
  &\textstyle-\frac{\sum_{(u,v)\in\mathcal P_{\mathit{cont}}}
  |c_{uv}^{\mathit{sim}}-c_{uv}^{\mathit{ref}}|}
  {|\mathcal P_{\mathit{cont}}|}\\[-1pt]
  &\textstyle-\frac{\sum_{a=1}^{2}\sum_{f\in\mathcal F}
  |s_{a,f}^{\mathit{sim}}-s_{a,f}^{\mathit{ref}}|}
  {2|\mathcal F|}
  \end{aligned}\right)$
& $1.0$ \\

\midrule

\multirow[t]{6}{=}{\textbf{Motion tracking}}
& Anchor position
& $\exp\!\left(
  -\frac{\|\mathbf p_{a,\mathit{anc}}^{\mathit{sim}}
  -\mathbf p_{a,\mathit{anc}}^{\mathit{ref}}\|_2^2}{0.3^2}
  \right)$
& $0.25$ \\[3pt]

& Anchor orientation
& $\exp\!\left(
  -\frac{d_Q(Q_{a,\mathit{anc}}^{\mathit{sim}},
  Q_{a,\mathit{anc}}^{\mathit{ref}})^2}{0.4^2}
  \right)$
& $0.25$ \\[3pt]

& Aligned body position
& $\exp\!\left(
  -\frac{\sum_{b\in\mathcal B_{\mathit{track}}}
  \|\mathbf p_{a,b}^{\mathit{sim}}
  -\widetilde{\mathbf p}_{a,b}^{\mathit{ref}}\|_2^2}
  {0.3^2\,|\mathcal B_{\mathit{track}}|}
  \right)$
& $0.5$ \\[3pt]

& Aligned body orientation
& $\exp\!\left(
  -\frac{\sum_{b\in\mathcal B_{\mathit{track}}}
  d_Q(Q_{a,b}^{\mathit{sim}},
  \widetilde Q_{a,b}^{\mathit{ref}})^2}
  {0.4^2\,|\mathcal B_{\mathit{track}}|}
  \right)$
& $0.5$ \\[3pt]

& Body linear velocity
& $\exp\!\left(
  -\frac{\sum_{b\in\mathcal B_{\mathit{track}}}
  \|\mathbf v_{a,b}^{\mathit{sim}}
  -\mathbf v_{a,b}^{\mathit{ref}}\|_2^2}
  {1.0^2\,|\mathcal B_{\mathit{track}}|}
  \right)$
& $0.5$ \\[3pt]

& Body angular velocity
& $\exp\!\left(
  -\frac{\sum_{b\in\mathcal B_{\mathit{track}}}
  \|\boldsymbol\omega_{a,b}^{\mathit{sim}}
  -\boldsymbol\omega_{a,b}^{\mathit{ref}}\|_2^2}
  {3.14^2\,|\mathcal B_{\mathit{track}}|}
  \right)$
& $0.5$ \\

\midrule

\multirow[t]{5}{=}{\textbf{Regularization}}
& Action rate
& $\sum_{a=1}^{2}
  \|\mathcal A_n^a-\mathcal A_{n-1}^a\|_2^2$
& $-0.3$ \\[3pt]

& Foot slip
& $\sum_{a=1}^{2}\sum_{f\in\mathcal F}
  s_{a,f}^{\mathit{sim}}
  \|\mathbf v_{a,f}^{\mathit{xy},\mathit{sim}}\|_2^2$
& $-0.5$ \\[3pt]

& Joint limits
& $\sum_{a=1}^{2}\sum_j
  \bigl([q_{a,j}^{\mathit{act}}-q_{a,j}^{\mathit{max}}]_+
  +[q_{a,j}^{\mathit{min}}-q_{a,j}^{\mathit{act}}]_+\bigr)$
& $-10.0$ \\[3pt]

& Undesired contacts
& $\sum_{a=1}^{2}\sum_{u\in\mathcal U}
  \mathbf 1\!\left[
  \max_{h\in\{0,1,2\}}\|\mathbf F_{a,u,h}^{\mathit{sim}}\|_2
  >1\,\mathrm N\right]$
& $-0.1$ \\[3pt]

& Torque
& $\sum_{a=1}^{2}\|\boldsymbol\tau_a\|_2^2$
& $-10^{-4}$ \\

\bottomrule
\end{tabularx}
}
\end{table}

\paragraph{Domain randomization.}
We randomize contact properties, torso center of mass, and default
joint offsets independently for the two humanoids.
At reset, both humanoids are initialized from a common reference
frame and receive the same sampled root-pose and root-velocity
perturbations, together with independent joint-position offsets.
We additionally randomize the delay of measured partner states,
using the same sampled delay in both communication directions
while leaving reference features undelayed.
These randomizations expose the residual expert to physical
parameter variation, initialization errors, and stale partner
information.
Table~\ref{tab:sipl_randomization} lists the ranges.
Continuous parameters are sampled uniformly within their intervals,
and communication delay is sampled uniformly from the listed
discrete values.

\begin{table}[t]
\centering
\caption{Domain randomization for residual interaction expert learning.}
\label{tab:sipl_randomization}
{\small
\setlength{\tabcolsep}{4pt}
\renewcommand{\arraystretch}{1.15}
\begin{tabularx}{\linewidth}{
@{}
>{\raggedright\arraybackslash}p{0.18\linewidth}
>{\raggedright\arraybackslash}p{0.28\linewidth}
>{\raggedright\arraybackslash}X
@{}
}
\toprule
\textbf{Category}
& \textbf{Parameter}
& \textbf{Range / setting} \\
\midrule

\multirow{5}{=}{Dynamics and contact}
& Static friction
& $[0.3,1.6]$ \\
& Dynamic friction
& $[0.3,1.2]$ \\
& Restitution
& $[0,0.5]$ \\
& Torso CoM offset
& $x:\ \pm0.025$\,m; $y,z:\ \pm0.05$\,m \\
& Default joint offset
& $\pm0.01$\,rad \\

\midrule

\multirow{5}{=}{Initial state}
& Root position offset
& $x,y:\ \pm0.05$\,m; $z:\ \pm0.01$\,m \\
& Root orientation offset
& Roll/pitch: $\pm0.1$\,rad; yaw: $\pm0.2$\,rad \\
& Root linear velocity offset
& $x,y:\ \pm0.5$\,m/s; $z:\ \pm0.2$\,m/s \\
& Root angular velocity offset
& Roll/pitch: $\pm0.52$\,rad/s; yaw: $\pm0.78$\,rad/s \\
& Joint position offset
& $\pm0.1$\,rad, clipped to soft joint limits \\

\midrule

Communication
& Peer measurement delay
& $\{0,1,2,3\}$ control steps ($0$--$60$\,ms) \\

\bottomrule
\end{tabularx}
}
\end{table}

\subsection{Flow-Matching Interaction Policy Learning}
\label{app:sipl_flow}

\paragraph{Shared policy and residual supervision.}
We consolidate the task-specific interaction experts into a shared
flow policy while retaining the frozen SONIC tracker.
The two humanoids share student parameters $\theta$, but each policy
evaluation produces only the ego humanoid's $29$-dimensional residual.
For task $k$ and humanoid $a$, the supervision target is the
deterministic correction from \redeqref{eq:interaction_residual}:
\begin{equation}
\Delta\mathcal A_{\mathit{expert}}^a
=
\Delta\mathcal A^{\bar a\rightarrow a}
\in\mathbb R^{29},
\qquad
a\in\{1,2\},\quad \bar a=3-a.
\label{eq:sipl_expert_target}
\end{equation}
Each target includes $\alpha_{\mathit{res}}$ but excludes exploration
noise.
Consistent joint ordering, action normalization, and the shared
frozen motor prior provide a common residual supervision space
across tasks.
Samples from both humanoids jointly update the same policy,
while supervision and prediction remain per-humanoid.

\paragraph{Ego-conditioned velocity network.}
Each policy evaluation organizes the paired observation windows
from the ego humanoid's perspective:
\begin{equation}
\hat{\mathcal O}^a
=
\left(
\mathcal O_{\mathit{his}}^a,
\mathcal O_{\mathit{fur}}^a,
\mathcal O_{\mathit{his}}^{\bar a},
\mathcal O_{\mathit{fur}}^{\bar a}
\right).
\label{eq:sipl_flow_condition}
\end{equation}
The observation groups follow Appendix~\ref{app:sipl_experts}.
The current partner-aware block $\mathcal O_{\mathit{ego}}^a$
is already contained in the history and is not introduced as an
additional token stream.
Separate history and future preprocessing MLPs, with weights
shared across roles, convert the four $20$-entry windows into
$80$ tokens.
Their representations retain temporal, stream, and ego/partner
identities.

A Transformer encodes these tokens, followed by global average
pooling and a linear projection to a conditioning embedding.
This embedding is concatenated with embeddings of the ego
humanoid's noisy residual and flow time.
An MLP then predicts the conditional velocity
$v_\theta(\Delta\mathcal A_t^a,t,\hat{\mathcal O}^a)
\in\mathbb R^{29}$.
The same network is evaluated for the other humanoid after
exchanging the ego and partner roles.
Thus, paired interaction context conditions each individual
correction without requiring a joint action-output head.
Table~\ref{tab:sipl_flow_settings} summarizes the architecture.

\paragraph{Student-state supervision.}
We alternate paired student rollouts and expert labeling using
DAgger-style supervision~\citep{ross2011reduction}.
At each visited interaction state, the shared policy is evaluated
for both humanoids, and each generated residual is added to the
corresponding frozen SONIC action.
Before the simulator advances, task $k$ selects the expert that
labels each humanoid using the same pre-action observations,
histories, and reference phase.
Expert parameters and normalization statistics remain fixed.

Each visited paired state therefore provides two training examples,
$(\hat{\mathcal O}^a,\Delta\mathcal A_{\mathit{expert}}^a)$,
one for each humanoid.
Both examples contribute to the same student update.
The labels specify corrections at student-visited states rather
than actions from separately executed expert trajectories.
Task identity routes the teacher but is not an input to the student.

\paragraph{Conditional flow matching.}
For each humanoid, we sample
$\boldsymbol\epsilon^a\sim\mathcal N(\mathbf0,\mathbf I_{29})$
and construct
\begin{equation}
\Delta\mathcal A_t^a
=
(1-t)\Delta\mathcal A_{\mathit{expert}}^a
+t\boldsymbol\epsilon^a,
\qquad t\in[0,1].
\label{eq:sipl_flow_path}
\end{equation}
Flow time $t$ is sampled from a Beta distribution and is distinct
from the physical control-step index $n$.
The endpoint $t=0$ corresponds to the expert residual and $t=1$
to Gaussian noise.
Applying \redeqref{eq:interaction_flow} to both humanoids yields
the shared-parameter objective
\begin{equation}
\mathcal L_{\mathit{fm}}(\theta)
=
\mathbb E\!\left[
\frac{1}{2}\sum_{a=1}^{2}
\left\|
v_\theta(\Delta\mathcal A_t^a,t,\hat{\mathcal O}^a)
-
\left(
\boldsymbol\epsilon^a
-\Delta\mathcal A_{\mathit{expert}}^a
\right)
\right\|_2^2
\right].
\label{eq:sipl_flow_shared_loss}
\end{equation}
The expectation covers labeled student states, flow times, and noise.
This objective jointly updates $\theta$ using both humanoids'
data, while each velocity prediction remains $29$-dimensional.
Only the flow policy is optimized; SONIC and the expert bank
remain frozen.
The updated shared policy then collects the next round of paired
interaction states.
At deployment, each humanoid runs a copy of the same trained
policy on top of its own SONIC tracker.
For humanoid $a$, we hold $\hat{\mathcal O}^a$ fixed, initialize
$\Delta\mathcal A_1^a$ with Gaussian noise, and integrate from
$t=1$ to $t=0$ using $N$ explicit Euler steps:
\begin{equation}
\Delta\mathcal A_{t_{j+1}}^a
=
\Delta\mathcal A_{t_j}^a
-\frac{1}{N}
v_\theta\!\left(
\Delta\mathcal A_{t_j}^a,t_j,\hat{\mathcal O}^a
\right),
\qquad
t_j=1-\frac{j}{N},
\label{eq:sipl_flow_euler}
\end{equation}
where $j=0,\ldots,N-1$.
The negative step traverses the learned expert-to-noise velocity
field in reverse.
Each policy evaluation generates only the ego humanoid's residual,
which is applied as
\begin{equation}
\mathcal A^a
=
\mathcal A_{\mathit{base}}^a
+\Delta\mathcal A_0^a.
\label{eq:sipl_flow_execution}
\end{equation}
No additional factor of $\alpha_{\mathit{res}}$ is applied because
the supervision targets already include this scale.
Independent deployment refers to separate policy evaluations and
action application, while partner information remains available
through the interaction-conditioning interface.
The shared flow policy therefore replaces the task-specific expert
bank while preserving the additive control interface used during
expert learning.

\begin{table}[t]
\centering
\caption{Architecture and learning specification of the shared
per-humanoid residual flow policy.}
\label{tab:sipl_flow_settings}

{\small
\setlength{\tabcolsep}{4pt}
\renewcommand{\arraystretch}{1.12}

\begin{tabularx}{\linewidth}{
@{}
>{\raggedright\arraybackslash}p{0.43\linewidth}
>{\raggedright\arraybackslash}X
@{}
}
\toprule
\textbf{Component}
& \textbf{Specification} \\
\midrule

\multicolumn{2}{c}{\textit{Architecture}} \\

History preprocessing MLP
& $239\rightarrow256\rightarrow256$; ReLU \\

Future preprocessing MLP
& $93\rightarrow256\rightarrow256$; ReLU \\

Preprocessing weight sharing
& Across roles; separate history/future weights \\

Token sequence per policy evaluation
& $2$ roles $\times2$ streams $\times20$ entries \\

Token width / count
& $256$ / $80$ \\

Token identity information
& Temporal position, stream, and ego/partner role \\

Condition encoder
& Transformer; global average pooling \\

Condition projection
& Linear; $1024$-dimensional output \\

Noisy residual embedding
& Linear $29\rightarrow1024$ \\

Flow-time embedding
& Sinusoidal; $1024$ dimensions \\

Feature fusion
& Concatenation; $3072$ dimensions \\

Velocity decoder hidden / output
& $[2048,2048,2048]$ / $29$ \\

Policy parameter sharing
& One parameter set for both humanoids \\

\midrule
\multicolumn{2}{c}{\textit{Learning and sampling}} \\

Training data
& Both humanoids' samples from paired rollouts \\

Teacher target
& Deterministic scaled $29$-D residual per humanoid \\

Task-ID conditioning
& Teacher routing only; excluded from student input \\

Trainable / frozen components
& Shared flow network / SONIC and residual experts \\

Noise distribution
& $\mathcal N(\mathbf0,\mathbf I_{29})$ per humanoid \\

Flow-time sampling
& Beta distribution on $[0,1]$ \\

Velocity target
& $\boldsymbol\epsilon^a
  -\Delta\mathcal A_{\mathit{expert}}^a$ \\

Loss aggregation
& Mean of both humanoids' flow-matching losses \\

Sampling direction / solver
& $t:1\rightarrow0$ / $N$ explicit Euler steps \\

Deployment
& Shared weights; separate per-humanoid inference \\

Action composition
& Each residual added to its own SONIC action \\

\bottomrule
\end{tabularx}
}
\end{table}

\paragraph{Real-world deployment.}
\label{app:deploy}
We deploy OmniInteract on two Unitree G1 humanoids
(approximately $1.3$~m tall, $29$ DoF), with each robot
independently executing the frozen SONIC tracker and the shared
residual flow policy onboard.
To construct partner-aware observations without privileged
simulator states, we employ a coarse-to-fine localization system
combining LiDAR and IMU measurements.
Point-LIO \citep{he2023point} provides LiDAR--inertial odometry at $10$~Hz.
For global initialization, GeoTransformer registers the current
scan against a pre-built point-cloud map, establishing a common
spatial frame for both robots.
During execution, a C++ relocalization node refines the pose
using Generalized Iterative Closest Point (GICP) \citep{segal2009generalized}, and a Kalman
filter fuses the odometry with global pose corrections to obtain
smooth state estimates and mitigate accumulated drift.
The robots exchange their estimated global poses and peer-state
observations through Lightweight Communications and Marshalling
(LCM) \citep{huang1998lcm}.
Received partner information is transformed into the ego-centric
frame and incorporated into the observation histories, while
the paired SCAR references advance with a common motion phase. 
The high-level policy are performed onboard. We utilize ONNX Runtime to execute the trained policies, achieving sufficiently low inference latency to sustain a stable 50~Hz control loop. Each robot constructs
its conditioning observations from the estimated states,
evaluates its frozen tracker, and integrates the flow policy
to obtain its own $29$-dimensional residual.
This correction is added to the base action following
\redeqref{eq:sipl_flow_execution}, and converted into desired joint
positions.
The low-level interface publishes motor targets and
proportional--derivative gains at $500$~Hz for joint-level
tracking.
\section{Experimental Details}
\label{app:exp}
\subsection{Interactoin Motion Benchmark}
\label{app:bench}
\paragraph{Interaction Motion Library} For retargeting, we evaluate on the full motion sets from the InterHuman \citep{liang2024intergen} and Inter-X \citep{xu2024inter} datasets. For control, we select a diverse subset of interaction
motions from both datasets, as summarized in Table~\ref{tab:interaction_motions}. 

\paragraph{Unseen Motion Set.}
In addition to interaction motions drawn from the training distributions, we
evaluate generalization on an unseen motion set. This set is constructed by
uniformly sampling 100 motion clips (10 seconds each) from each of the
retargeted InterHuman and Inter-X datasets, yielding 200 clips in total, while
explicitly excluding all motions used for residual-expert training or unified
policy distillation. The unseen set covers diverse interaction patterns and
contact configurations, and is used to assess generalization beyond the
curated interaction motion library.

\begin{table*}[!t]
\centering
\caption{Summary of interaction motions in the control benchmark.}
\label{tab:interaction_motions}

{\fontsize{7.5}{9}\selectfont
\setlength{\tabcolsep}{2pt}
\renewcommand{\arraystretch}{0.88}

\begin{tabularx}{\textwidth}{
@{}
>{\raggedright\arraybackslash}p{0.1\textwidth}
>{\raggedright\arraybackslash}p{0.230\textwidth}
>{\raggedright\arraybackslash}p{0.075\textwidth}
>{\raggedright\arraybackslash}p{0.100\textwidth}
>{\raggedright\arraybackslash}X
@{}
}
\toprule
Motion ID
& Motion Source
& Duration
& Category
& Motion Description \\
\midrule

\multicolumn{5}{@{}l@{}}{\cellcolor[gray]{0.93}\textbf{\textit{InterHuman}}} \\

1  & InterHuman-2312 & 8.0 s  & Dance
& Synchronized dance with shoulder and waist contact. \\

2  & InterHuman-2853 & 58.4 s & Push
& Repeated right-hand pushing, blocking. \\

3  & InterHuman-6436 & 29.6 s & Dance
& Two people dance together. \\

4  & InterHuman-6494 & 23.9 s & Embrace
& Two people dance while embracing. \\

5  & InterHuman-6774 & 32.3 s & Dance
& Face-to-face ballroom dance. \\

6  & InterHuman-6791 & 25.5 s & Dance
& Collaborative dance after approaching with raised arms. \\

7  & InterHuman-2206 & 22.1 s & Embrace
& Close embrace, swaying, walking, and a guided turn. \\

8  & InterHuman-2239 & 57.0 s & Dance
& Dancing followed by hugging and spinning. \\

9  & InterHuman-2670 & 77.2 s & Spin
& Hand-clasped spinning followed by an embrace. \\

10 & InterHuman-2861 & 60.4 s & TaiChi
& Partner tai chi with continuous hand contact. \\

11 & InterHuman-2916 & 53.9 s & Push
& Alternating left- and right-hand pushing and defending. \\

12 & InterHuman-2924 & 85.7 s & Combat
& Repeated attacks, defense, and counter-pushing. \\

13 & InterHuman-2932 & 83.7 s & Struggle
& Bilateral hand-contact struggle with pushing and defense. \\

14 & InterHuman-3857 & 44.8 s & Sparring
& Martial-arts sparring between two people. \\

15 & InterHuman-6391 & 20.3 s & Sway
& Handholding with repeated side-to-side swaying. \\

16 & InterHuman-6809 & 23.4 s & Dance
& Contact dance with circular spinning. \\

17 & InterHuman-6893 & 71.8 s & Embrace
& Approaching, embracing, and swaying together. \\

18 & InterHuman-2115 & 10.6 s & Stretch
& Alternating two-handed grasp and forward arm stretch. \\

19 & InterHuman-2116 & 12.1 s & Spin
& Handholding, raised arms, and turning. \\

20 & InterHuman-2251 & 19.9 s & Embrace
& Embracing while moving in a circle. \\

21 & InterHuman-2736 & 10.2 s & Conversation
& Laughing and conversing face to face. \\

22 & InterHuman-2851 & 63.5 s & TaiChi
& Partner tai chi practice. \\

23 & InterHuman-319 & 10.7 s & Grasp
& Approaching quickly, grasping a hand, and bowing. \\

24 & InterHuman-327 & 12.0 s & Handshake
& Approaching, shaking hands, and stepping backward. \\

25 & InterHuman-7023 & 16.5 s & Wrestling
& Argument followed by wrestling. \\

26 & InterHuman-7207 & 14.2 s & Grapple
& Argument, shoving, and arm grappling. \\

27 & InterHuman-7335 & 10.0 s & Care
& Bandaging a partner's hand followed by a hug. \\

28 & InterHuman-7619 & 10.0 s & Snatch
& Snatching an object followed by a physical struggle. \\

29 & InterHuman-1557 & 61.7 s & Boxing
& Boxing with repeated exchanges of punches. \\

30 & InterHuman-785 & 9.6 s & Fencing
& Two-person fencing practice. \\

\multicolumn{5}{@{}l@{}}{\cellcolor[gray]{0.93}\textbf{\textit{Inter-X}}} \\

31 & Inter-X-G013T005A021R003 & 23.3 s & Dance
& Dancing and spinning while holding both hands. \\

32 & Inter-X-G014T004A021R005 & 21.4 s & Dance
& Face-to-face partner dance with arm and hand contact. \\

33 & Inter-X-G017T000A000R002 & 20.9 s & Hug
& A hesitant embrace followed by reciprocation. \\

34 & Inter-X-G017T004A021R007 & 25.5 s & Dance
& One partner guides the other's arms while dancing. \\

35 & Inter-X-G031T005A021R005 & 23.5 s & Dance
& Partner swaying with shoulder and waist contact. \\

36 & Inter-X-G053T000A001R011 & 27.7 s & Handshake
& Face-to-face two-handed handshake. \\

37 & Inter-X-G055T003A016R008 & 27.0 s & High-five
& Playful sequence of stepping on each other's feet. \\

38 & Inter-X-G057T000A000R004 & 20.4 s & Hug
& Side embrace around the partner's waist. \\

39 & Inter-X-G004T008A021R005 & 20.1 s & Dance
& Face-to-face dance with side jumps and kicks. \\

40 & Inter-X-G007T001A006R002 & 21.7 s & Posing
& Coordinated hand gestures and paired posing. \\

41 & Inter-X-G024T005A028R019 & 20.3 s & Wrestling
& Standing hand wrestling with crossed wrists. \\

42 & Inter-X-G028T004A021R005 & 57.4 s & Dance
& Close partner dance with waist and shoulder contact. \\

43 & Inter-X-G028T004A021R010 & 25.3 s & Dance
& Partner dance with handholding and raised arms. \\

44 & Inter-X-G029T003A016R009 & 59.5 s & High-five
& Mutual foot-stepping followed by a high-five. \\

45 & Inter-X-G031T005A021R006 & 27.3 s & Dance
& Partner dance with waist contact and handholding. \\

46 & Inter-X-G034T003A012R012 & 20.2 s & Pointing
& Pointing at the partner and then behind them. \\

47 & Inter-X-G034T007A036R008 & 21.9 s & Covering
& Reaching out to cover the partner's mouth. \\

48 & Inter-X-G052T004A022R011 & 22.2 s & Linking
& Walking side by side with linked arms. \\

49 & Inter-X-G053T005A021R007 & 37.6 s & Dance
& Handholding and circular partner dance. \\

50 & Inter-X-G055T001A008R019 & 23.7 s & Pull
& Grasping and pulling the partner closer. \\

51 & Inter-X-G034T004A017R013 & 25.2 s & Chase
& Taking turns chasing and evading each other. \\

52 & Inter-X-G053T000A003R022 & 12.1 s & Grab
& Grabbing the partner's hand and pulling them closer. \\

53 & Inter-X-G053T000A004R027 & 15.2 s & Hit
& Striking the partner's shoulder with a fist. \\

54 & Inter-X-G009T006A033R009 & 42.6 s & Imitation
& Imitating gestures, steps, and leg movements. \\

55 & Inter-X-G016T001A005R001 & 13.6 s & Kick
& Kicking toward a partner's leg while they dodge. \\

56 & Inter-X-G028T002A014R026 & 12.4 s & Collision
& Bumping the partner's shoulder with one's own. \\

57 & Inter-X-G052T006A030R000 & 15.3 s & Patting
& Approaching from the side and patting a cheek. \\

58 & Inter-X-G016T004A020R001 & 20.7 s & Game
& Playing rock-paper-scissors face to face. \\

59 & Inter-X-G017T002A010R002 & 12.5 s & Slap
& Exchanging slaps to the sides of the face. \\

60 & Inter-X-G053T003A018R017 & 14.9 s & Whisper
& Approaching and whispering near a partner's ear. \\

\bottomrule
\end{tabularx}
}
\end{table*}

\subsection{Baseline Implementation Details}
\label{app:baseline}
To validate our contributions, we benchmark our framework against two sets of baselines:  retargeting methods (answering Q1) and control policy variants (answering Q2 \& Q3).
\paragraph{Retargeting Baselines.} For fair comparison, all baselines use the same source motions and identical skeletal-scaling preprocessing.

\begin{itemize}[]
\item \textbf{GMR (General Motion Retargeting)} \citep{araujo2025retargeting}: A Cartesian-space optimization-based retargeting method that independently fits each humanoid to the source motion by minimizing positional and rotational errors of selected \textbf{body keypoints}, such as the pelvis and end-effectors. Since the two agents are optimized separately, GMR does not explicitly model inter-agent interaction constraints.

\item \textbf{OmniRetarget} \citep{yang2025omniretarget}: A topology-aware retargeting method that uses an \textbf{Interaction Mesh} (I-Mesh) to preserve geometric structure, originally developed for human–object interaction. In our dual-humanoid setting, we apply OmniRetarget independently to each agent. As a result, it preserves per-agent body geometry but does not explicitly constrain the relative spatial relationships between the two humanoids.

\item \textbf{PHC} \citep{luo2023perpetual}: A \textbf{sparse correspondence-based} humanoid retargeting pipeline that fits predefined human joints to robot links. We apply it independently to each agent, preserving individual motion but without explicit inter-agent geometric constraints.

\item \textbf{UMR (Unified Motion Retargeting)} \citep{cao2026unified}:A dense surface correspondence-based retargeting method that aligns human and robot exterior \textbf{point clouds} through constrained optimization. We apply it independently to each agent, providing fine-grained body-surface alignment but without explicit pairwise constraints between the two humanoids.

\end{itemize}

\paragraph{Control Policy Baselines.} We evaluate our scalable flow-based interaction policy training framework by comparing against representative single-agent baselines and ablated variants. All variants are trained using the same PPO hyperparameters.

\begin{itemize}[]
\item \textbf{TWIST2} \citep{ze2025twist2}: An RL-based whole-body motion tracker conditioned on reference motion and
robot proprioception. We initialize from the official checkpoint and adapt
the tracker on our interaction-motion subsets, applying it independently to
each humanoid with its own reference motion. The policy receives neither
partner-state information nor explicit inter-humanoid interaction cues.

\item \textbf{BeyondMimic} \citep{liao2026beyondmimic}: An RL-based motion-tracking framework that trains a dedicated policy for each
reference motion, targeting high-dynamic whole-body tracking. In our
dual-humanoid setting, we train one tracker per motion and execute the two
corresponding policies independently for the two robots.

\item \textbf{SONIC} \citep{luo2026sonic}: A large-scale universal whole-body motion tracker that uses a single unified policy to track diverse reference motions, including unseen motions. We independently run the same tracker for both humanoids with their respective references, without explicit peer-state or interaction-aware constraints.

\item \textbf{From scratch RL (w/o Residual Learning)}: Evaluates the contribution of residual interaction learning built upon a pretrained single-agent motion prior. \textit{Implementation}: The frozen base tracker and residual decomposition are removed, and a peer-aware policy is trained from scratch with PPO to directly predict the full control actions for both humanoids. The observation space and interaction rewards remain unchanged.

\item \textbf{Specialist$\rightarrow$Unified MLP (w/o Flow Matching)}: Evaluates the contribution of flow-based policy consolidation when integrating interaction specialists into a unified controller. 
\textit{Implementation}: The residual specialists and DAgger data-collection procedure remain unchanged, while the conditional flow-matching policy is replaced with a deterministic MLP that directly regresses the paired residual actions from the same conditioning inputs and expert targets.
\end{itemize}

\subsection{Evaluation Protocol and Metric Details}
\label{app:metrics}
This subsection details the evaluation protocol and metrics used throughout the paper. We use task-specific metrics to evaluate retargeting quality and dynamic control
performance of our framework, respectively.

\noindent\textit{1) Evaluation Protocol}
\paragraph{Motion segmentation and episode length.} For control policy evaluation, to ensure a consistent evaluation horizon, we divide all reference motions into
10-second clips, corresponding to 500 control steps and matching the episode
length used during training. Motions shorter than 10 seconds are evaluated in
their original duration. Each motion clip is evaluated over 10 independent rollouts, and metrics are averaged across these trials.

\noindent\textit{2) Metrics Details}
\paragraph{Retargeting Evaluation Metrics.}
We evaluate retargeting quality from three complementary aspects:
physical feasibility, interaction fidelity, and downstream learnability.

\begin{itemize}

\item \textbf{Maximum Penetration Depth (MPD).}
Measures physical plausibility by reporting the maximum inter-humanoid
penetration depth over each motion sequence. Penetration is computed using
the MuJoCo~\citep{todorov2012mujoco} collision proxy geometries. We first
compute MPD for each sequence and then average it over the evaluation set.
Lower values indicate fewer severe inter-body collisions.

\item \textbf{Contact Interaction Error (CIE).}
Measures fine-grained contact fidelity using cross-body surface pairs whose
distance in the source motion is within the contact threshold
$\tau_c=2\,\mathrm{cm}$. For each pair, we compute the $L_2$ deviation
between the source and retargeted 3D relative vectors and normalize it by the
canonical humanoid body scale $s_{\mathrm{body}}=1.32\,\mathrm{m}$.
CIE is reported as a percentage, with lower values indicating better
preservation of contact distance and direction.

\item \textbf{Near-field Geometry Error (NGE).}
Measures preservation of the surrounding interaction geometry using
cross-body surface pairs whose source distances lie between
$\tau_c=2\,\mathrm{cm}$ and $\tau_n=15\,\mathrm{cm}$. Pairs whose two
endpoints are both contact points are excluded. The $L_2$ deviation between
the source and retargeted relative vectors is normalized by
$s_{\mathrm{body}}$ and reported as a percentage. Lower NGE indicates better
preservation of near-field spatial structure.

\item \textbf{Contact F1 (C-F1).}
Measures region-level contact preservation. A body-region pair is considered
in contact if at least one cross-body surface pair lies within
$\tau_c=2\,\mathrm{cm}$. We compare the source and retargeted contact labels
and report the micro-averaged F1 score over all frames as a percentage.

\item \textbf{Downstream Success Rate (DSR).}
Serves as a proxy for downstream physical learnability. We define DSR as the
percentage of RL rollouts that successfully execute the retargeted reference
while preserving the interaction structure, requiring
$\mathrm{CIE}<20\%$ and $\mathrm{NGE}<20\%$ relative to the scaled reference.

\end{itemize}

\paragraph{Control Policy Evaluation Metrics.}
We evaluate dual-humanoid control in terms of motion tracking,
interaction preservation, and overall execution success with respect to the
retargeted reference trajectories.

\begin{itemize}[]

\item \textbf{Root-relative Mean Per-Joint Position Error
($E_{\textit{mpjpe}}$).}
Measures individual motion-tracking fidelity as the mean per-joint position
error after removing global root translation. We compute
$E_{\textit{mpjpe}}$ over both humanoids and average it over the rollout.
Lower values indicate more accurate whole-body tracking independent of global
translational drift.

\item \textbf{Interaction Geometry Error
($E_{\textit{inter-geo}}$).}
Measures how faithfully the executed motion preserves the contact
interfaces and surrounding interaction geometry of the retargeted
reference. We reuse the CIE and NGE definitions above, replacing the
source--retargeted comparison with a reference--execution comparison.
Contact and near-field surface pairs are determined exclusively from
the retargeted reference using the same thresholds and exclusion rules.
For each component, relative-vector deviations are averaged over all
corresponding frame--pair instances in the rollout, normalized by
$s_{\mathrm{body}}$ and reported as a percentage.

\item \textbf{Success Rate (Succ.).}
Measures overall interaction execution success. A rollout is considered
successful when $E_{\textit{mpjpe}}<10\,\mathrm{cm}$,
$E_{\textit{inter-geo}}<20\%$, and neither humanoid falls or terminates
early. We report the percentage of successful rollouts over all evaluation
trials.

\end{itemize}

\section{Additional Experiments}
\label{app:additional}
\subsection{Ablation Study \& Additional Visualization on SCAR}

\begin{table}[H]
\centering
\caption{\textbf{SCAR component ablation on Inter-X and InterHuman.}
All metrics are reported in \%; arrows indicate the preferred direction,
and the best results are in bold.}
\label{tab:scar_component_ablation}
\vspace{0.2em}
\begingroup
\small
\setlength{\tabcolsep}{5pt}
\renewcommand{\arraystretch}{1.12}
\begin{tabular*}{0.96\linewidth}{@{\extracolsep{\fill}}lcccccc@{}}
\toprule
\multirow{2}{*}{\textbf{Variant}}
 & \multicolumn{3}{c}{\textbf{Inter-X}}
 & \multicolumn{3}{c}{\textbf{InterHuman}} \\
\cmidrule(lr){2-4}\cmidrule(lr){5-7}
 & CIE $\downarrow$ & NGE $\downarrow$ & DSR $\uparrow$
 & CIE $\downarrow$ & NGE $\downarrow$ & DSR $\uparrow$ \\
\midrule
w/o CIO & 6.78 & 5.43 & 64.2 & 7.45 & 7.21 & 68.8 \\
w/o NCO & 4.21 & 6.63 & 65.8 & 5.23 & 8.21 & 69.2 \\
\midrule
\textbf{SCAR (full)}
 & \textbf{3.98} & \textbf{4.99} & \textbf{72.1}
 & \textbf{4.96} & \textbf{6.49} & \textbf{75.4} \\
\bottomrule
\end{tabular*}
\endgroup
\end{table}

\begin{figure}[H]
    \centering
    \includegraphics[width=0.85\linewidth]{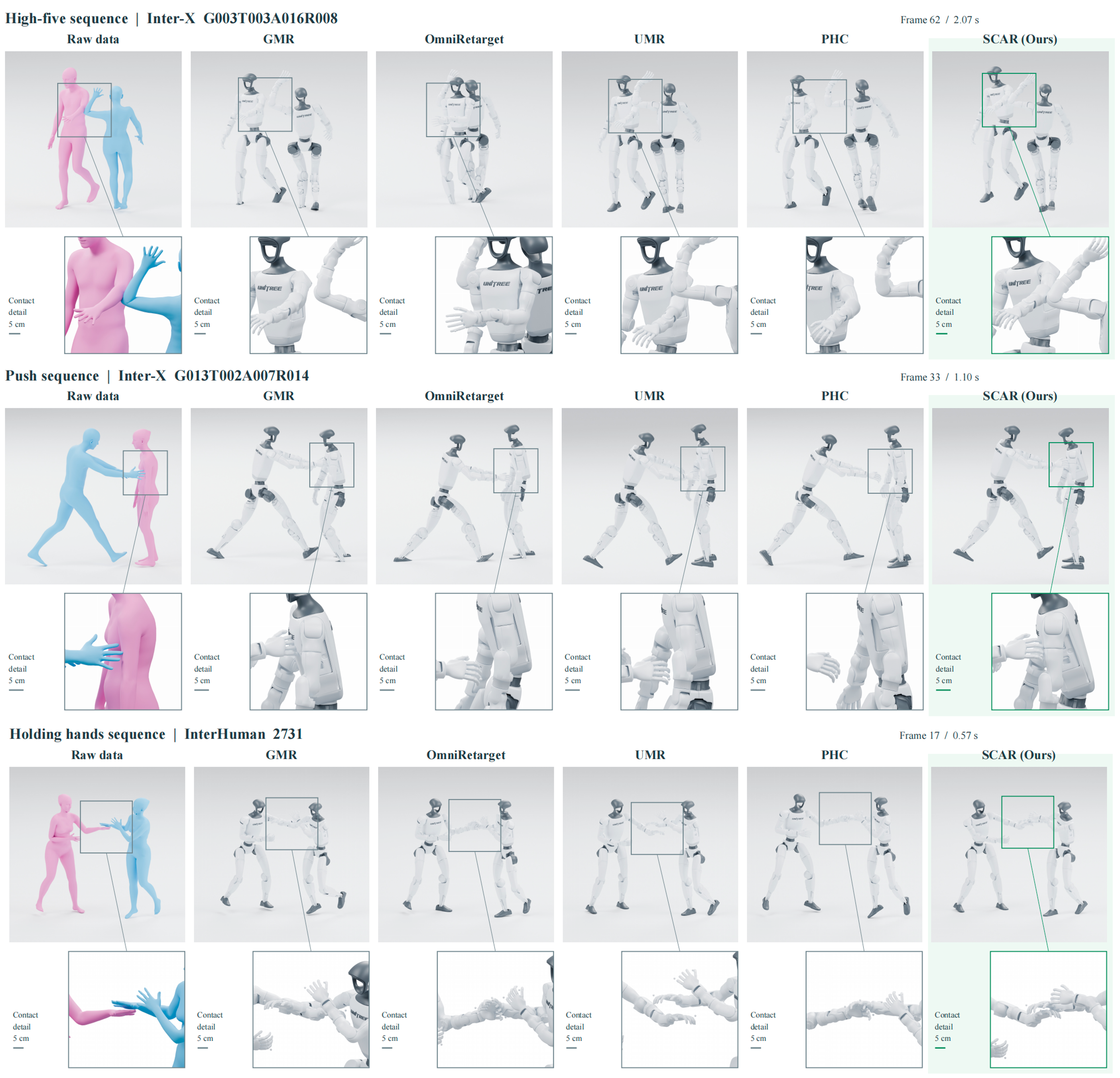}
\caption{
Qualitative comparison of interaction retargeting on Inter-X and InterHuman. (Top: High-five, Middle: Push, Bottom: Holding Hands.) Columns show the raw motion and results from GMR, OmniRetarget, UMR, PHC, and SCAR (ours). Zoomed-in views highlight contact details.
}
    \label{app2}
\end{figure}

\paragraph{Component ablation.}
As shown in Table~\ref{tab:scar_component_ablation}, full SCAR
achieves the best results across all metrics on both datasets.
Removing \textbf{Contact-Interface Objective (CIO)} primarily increases CIE, from 3.98\% to 6.78\%
on Inter-X and from 4.96\% to 7.45\% on InterHuman.
Removing \textbf{Near-Field Context Objective (NCO)} instead leads to a larger increase in NGE,
from 4.99\% to 6.63\% and from 6.49\% to 8.21\%, respectively.
Both ablations also reduce DSR, suggesting that CIO and NCO
play complementary roles in preserving interaction quality
and supporting successful motion execution.

\paragraph{Qualitative comparison.}
Figure~\ref{app2} provides additional comparisons on high-five,
push, and holding-hands interactions.
The enlarged views highlight differences in the relative
placement of interacting body parts.
Compared with the baselines, SCAR more closely preserves
the contact relationships in the source motions, particularly
hand-to-body alignment in pushing and hand-to-hand alignment
in holding hands. These examples qualitatively support
SCAR's effectiveness in preserving interaction-specific
spatial relationships during retargeting.

\subsection{Sensitivity Analysis and Qualitative Results of OmniInteract}
\label{exp:real}

We further examine the sensitivity of OmniInteract to the residual action scale and provide qualitative examples of interaction retargeting in simulation and humanoid interactions in the real world.

\begin{figure}[H]
    \centering
    \includegraphics[width=\linewidth]{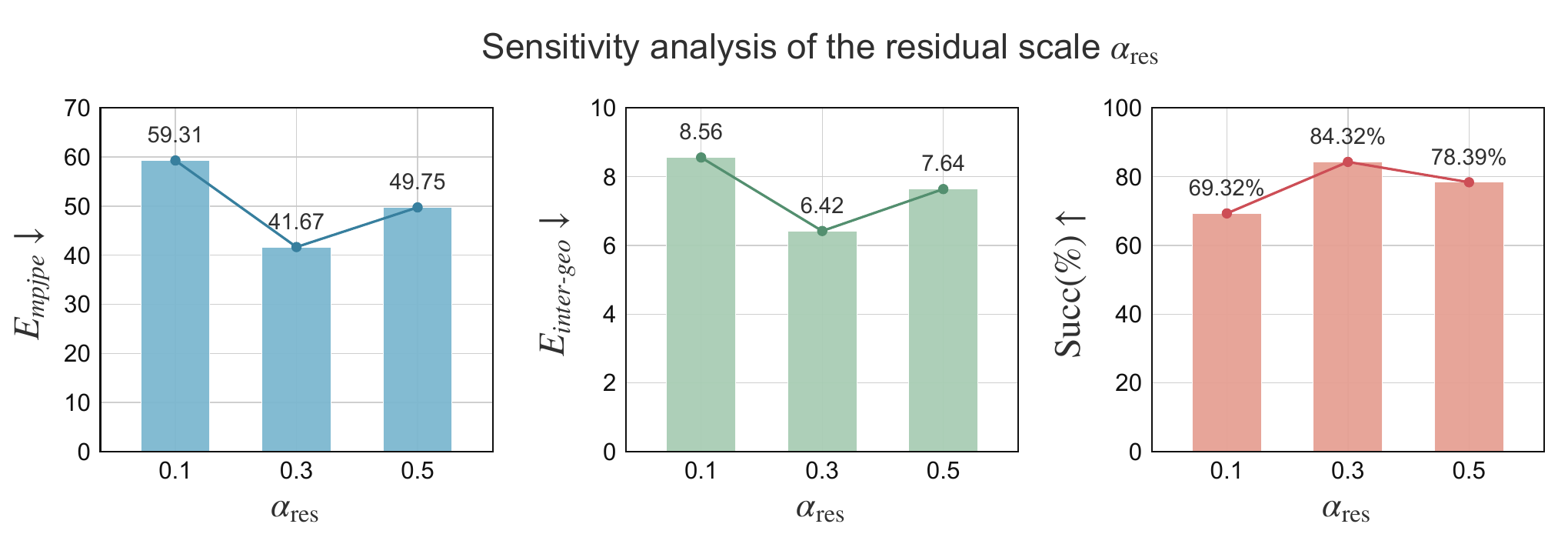}
\caption{
Sensitivity to the residual action scale \(\alpha_{\mathrm{res}}\). A scale of 0.3 achieves the lowest tracking and interaction geometry errors and the highest success rate.
}
    \label{app5}
\end{figure}

Figure~\ref{app5} evaluates the residual action scale $\alpha_{\mathrm{res}} \in \{0.1, 0.3, 0.5\}$. Among the tested values, $\alpha_{\mathrm{res}}=0.3$ achieves the best performance across all three metrics, with $E_{mpjpe}=41.67$, $E_{inter\text{-}geo}=6.42$, and a success rate of $84.32\%$. Both smaller and larger scales increase the errors and reduce the success rate, highlighting the importance of appropriately scaling the residual correction.

\begin{figure}[H]
    \centering
    \includegraphics[width=\linewidth]{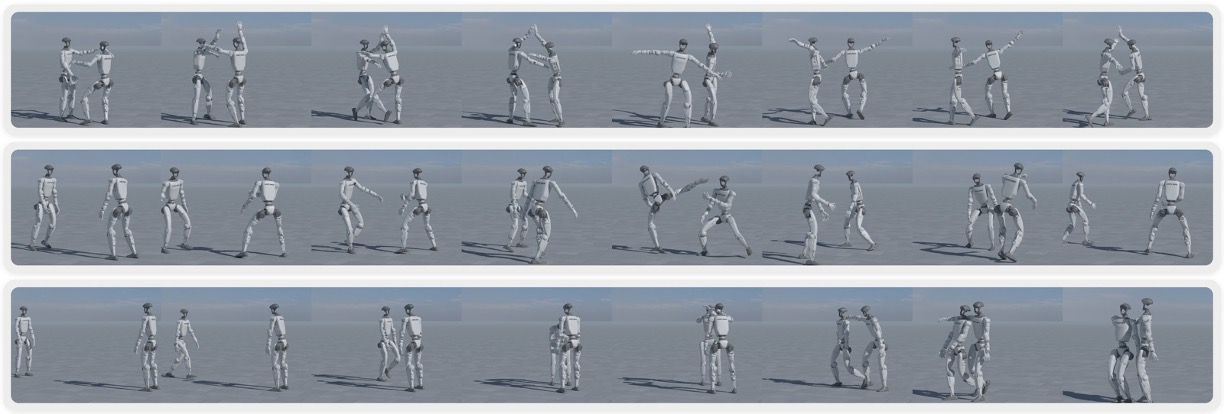}
\caption{
Qualitative results of OmniInteract for interaction control in Isaac simulation. From top to bottom: dancing, fighting, and walking with an arm around a partner’s shoulder.
}
    \label{app3}
\end{figure}

Figure~\ref{app3} visualizes the results of OmniInteract for interaction control in Isaac simulation, including dancing, fighting, and walking with an arm around a partner's shoulder. These examples complement the quantitative evaluation by illustrating the robot poses and relative spatial configurations associated with different interaction patterns.

\begin{figure}[H]
    \centering
    \includegraphics[width=\linewidth]{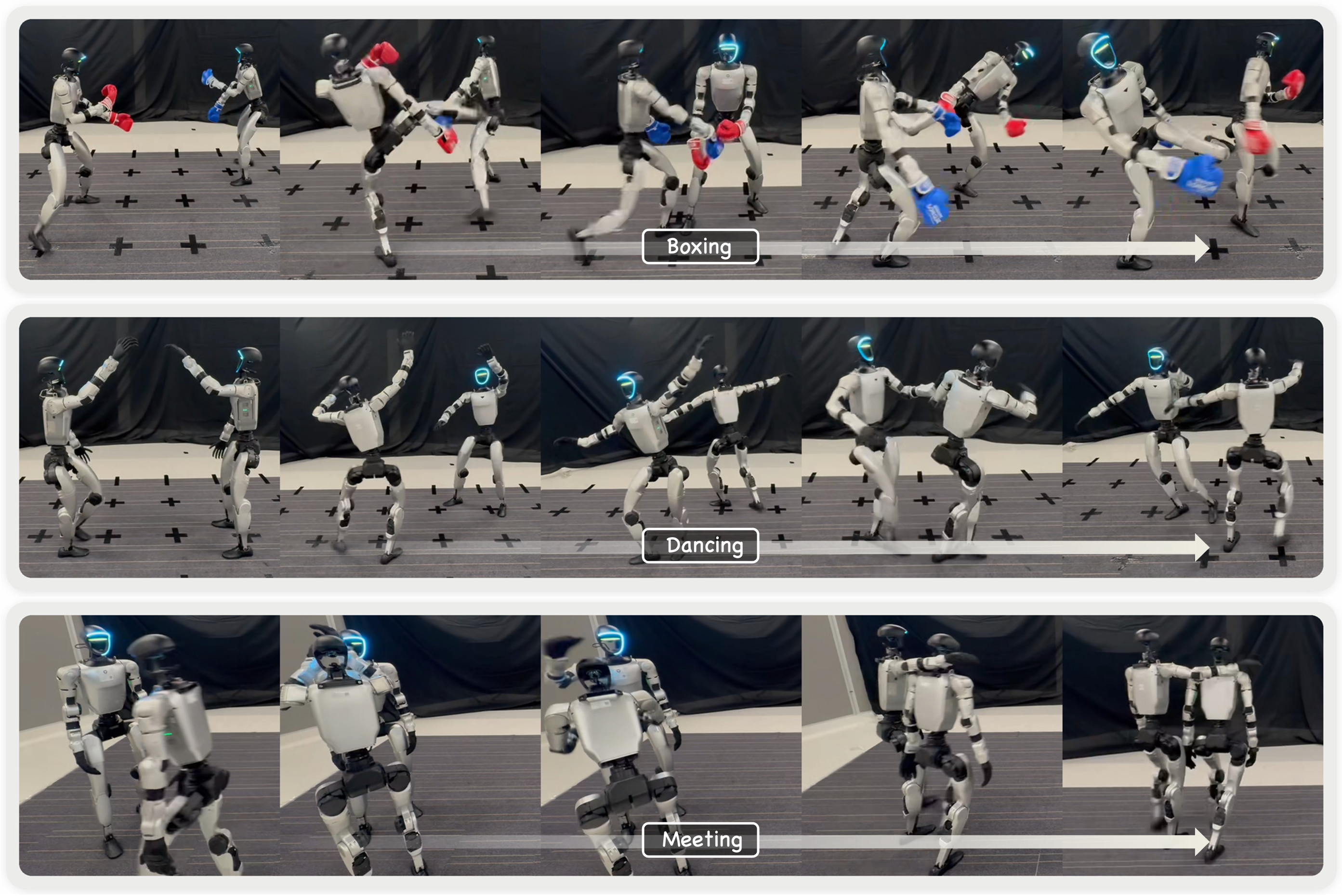}
\caption{
Qualitative visualization of real-world humanoid interactions. (Top: Boxing, Middle: Dancing, Bottom: Meeting.) Each row presents five keyframes arranged chronologically from left to right, illustrating coordinated movements and close-range interactions between two humanoid robots.)
}
    \label{app4}
\end{figure}

Figure~\ref{app4} presents additional real-world examples of boxing, dancing, and meeting. Each row contains five keyframes ordered chronologically from left to right, showing how the two humanoid robots' poses and relative positions evolve during execution. Together, these examples illustrate the range of coordinated and close-range interactions considered in our evaluation.

\end{document}